\documentclass[lettersize,journal]{IEEEtran}
\usepackage{amsmath,amsfonts}
\usepackage{array}
\usepackage{textcomp}
\usepackage{url}
\usepackage{verbatim}
\usepackage{graphicx}
\def\BibTeX{{\rm B\kern-.05em{\sc i\kern-.025em b}\kern-.08em
    T\kern-.1667em\lower.7ex\hbox{E}\kern-.125emX}}
\usepackage{balance}
\usepackage{marvosym}

\usepackage{bm}
\usepackage[ruled,linesnumbered]{algorithm2e}
\usepackage[x11names]{xcolor}
\usepackage[breaklinks=true,colorlinks,bookmarks=false,citecolor=SpringGreen4,urlcolor=black]{hyperref}
\usepackage{booktabs}
\usepackage{subcaption}
\usepackage{colortbl}
\usepackage{multirow}
\usepackage{framed}
\usepackage{scalefnt}
\usepackage{pifont}
\usepackage{arydshln}

\newcommand{\cmark}{\ding{51}}%
\newcommand{\xmark}{\ding{55}}%

\newcommand{\etal}{\textit{et al}.}
\newcommand{\ie}{\textit{i}.\textit{e}.}
\newcommand{\eg}{\textit{e}.\textit{g}.}
\def\etc{\emph{etc}}

\definecolor{Gray1}{gray}{0.8}
\definecolor{Gray2}{gray}{0.9}

\begin{document}
\title{CLIP4STR: A Simple Baseline for Scene Text Recognition with Pre-trained Vision-Language Model}

\author{Shuai Zhao, Ruijie Quan\textsuperscript{\Letter}, Linchao Zhu, Yi Yang~\IEEEmembership{Senior~Member,~IEEE}
\thanks{This work was partially supported by the Earth System Big Data Platform of the School of Earth Sciences, Zhejiang University. Corresponding author: Ruijie Quan}

\thanks{
Shuai Zhao is with the ReLER Lab, Australian Artificial
Intelligence Institute, University of Technology Sydney, Ultimo, NSW 2007, Australia. Part of this work is done during an internship at Baidu Inc. E-mail: zhaoshuaimcc@gmail.com. Linchao Zhu, Ruijie Quan, Yi Yang are with ReLER Lab, CCAI, Zhejiang University, Zhejiang, China. E-mail: \{zhulinchao, quanruijie, yangyics\}@zju.edu.cn.}
}


\maketitle

\begin{abstract}
Pre-trained vision-language models~(VLMs) are the de-facto foundation models for various downstream tasks.
However, scene text recognition methods still prefer backbones pre-trained on a single modality, namely, the visual modality, despite the potential of VLMs to serve as powerful scene text readers.
For example, CLIP can robustly identify regular (horizontal) and irregular (rotated, curved, blurred, or occluded) text in images.
With such merits, we transform CLIP into a scene text reader and introduce CLIP4STR,
a simple yet effective STR method
built upon image and text encoders of CLIP.
It has two encoder-decoder
branches: a visual branch and
a cross-modal branch.
The visual branch provides an initial prediction based on the visual feature,
and the cross-modal branch refines this prediction by addressing the discrepancy between the visual feature and text semantics.
To fully leverage the capabilities of both branches,
we design a dual predict-and-refine decoding scheme for inference.
We scale CLIP4STR in terms of the model size, pre-training data, and training data, achieving state-of-the-art performance on 13 STR benchmarks.
Additionally, a comprehensive empirical study
is provided to enhance the understanding of the adaptation of CLIP to STR.
Our method establishes a simple yet strong baseline for future STR research with VLMs.
\end{abstract}



\section{Introduction} \label{sec-intro}
\IEEEPARstart{V}{ision-language} models~(VLMs)
pre-trained on web-scale data
like CLIP~\cite{2021-clip} and ALIGN~\cite{2021-align} shows
remarkable zero-shot capacity across different tasks.
Researchers also successfully transfer the knowledge from pre-trained VLMs to diverse tasks in a zero-shot or fine-tuning manner,
\eg, visual question answering~\cite{2022_haoyu_clip},
information retrieval~\cite{2022_clip4clip,yang2024dgl},
referring expression comprehension~\cite{2022_reclip},
and image captioning~\cite{2021_CLIPScore}.
VLM is widely recognized as a foundational model and an important component of artificial intelligence~\cite{fei2022towards}.

\begin{figure}[!t]
	\centering
	\includegraphics[width=0.9\linewidth]{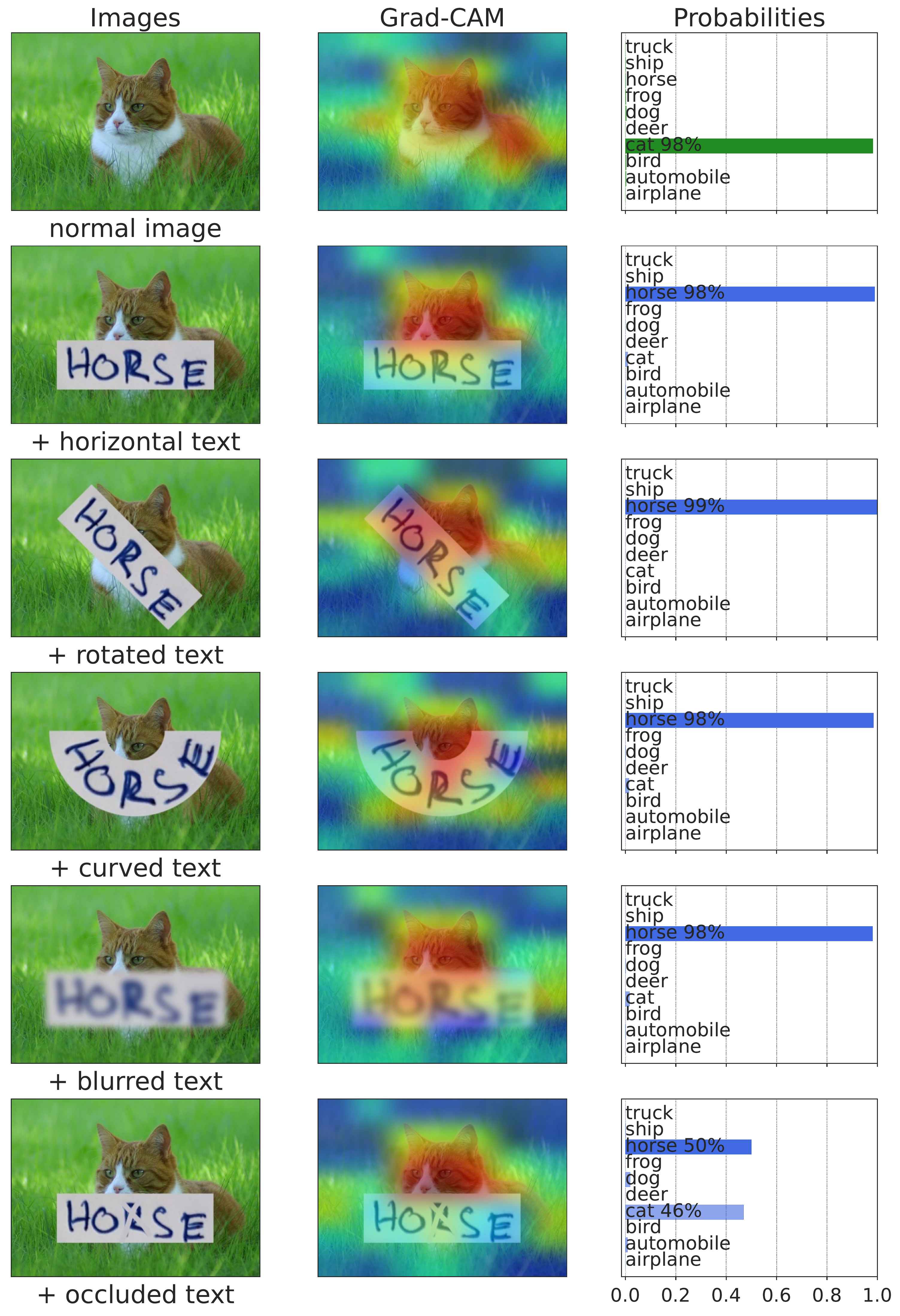}
	\caption{\textbf{Zero-shot classification results of CLIP-ViT-B/32}. CLIP can perceive and understand text in images, even for irregular text with noise, rotation, and occlusion.
 CLIP is potentially a powerful scene text recognition expert.}
	\label{fig:clip_0shot}
\end{figure}

Scene text recognition (STR) is a critical technique and an essential process in many vision and language applications, \eg, document analysis,
autonomous driving, and augmented reality.
Similar to the aforementioned cross-modal tasks,
STR involves two different modalities: image and text.
However, unlike the popularity of pre-trained VLMs in other cross-modal tasks, STR methods still tend to rely on backbones pre-trained on single-modality data~\cite{2021_abinet,atienza2021vision,yang2022reading,li2021trocr}.
In this work, we show that VLM pre-trained on image-text pairs possess strong scene text perception abilities, making them superior choices as STR backbones.

\begin{figure}[!t]
	\centering
	\includegraphics[width=0.66\linewidth]{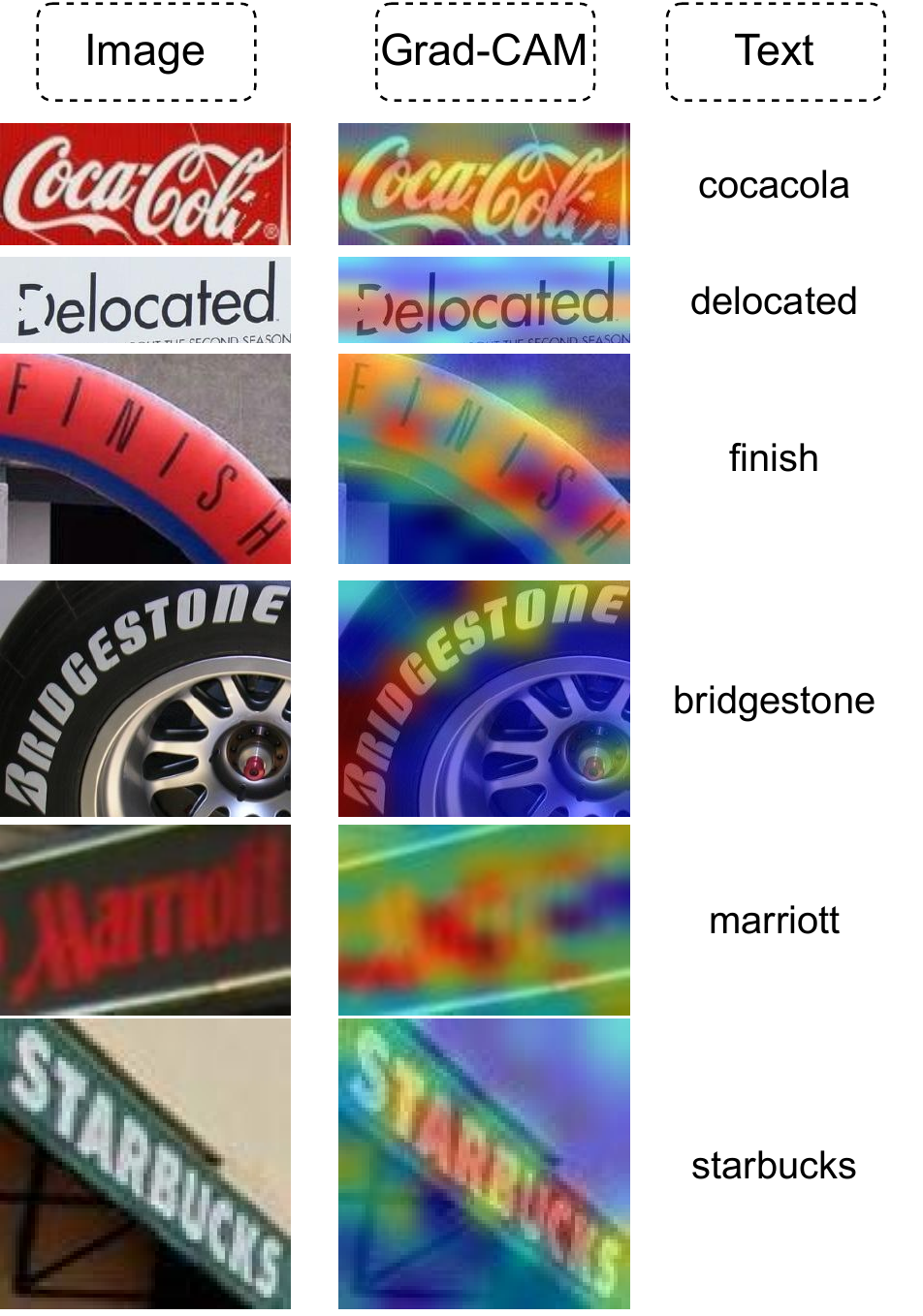}
	\caption{\textbf{Attention of CLIP-ViT-B/32 for STR images}.}
	\label{fig:str-cam}
\end{figure}

STR methods generally struggle with irregular text like rotated, curved, blurred, or occluded text~\cite{2021_long_survey,raisi2022occluded}.
However, irregular text is prevalent in real-life scenarios~\cite{2019_icdar_art,zhang2017uber},
making it necessary for STR models to effectively handle these challenging cases.
Interestingly, we observe that the VLM~(\eg, CLIP~\cite{2021-clip}) can robustly perceive irregular text in natural images.
In Fig.~\ref{fig:clip_0shot}, we put different text 
stickers on a natural image, use CLIP to classify it\footnote{The class categories are from CIFAR-10~\cite{krizhevsky2009learning}.
The experiment is inspired by Stanislav Fort~\cite{Fort2021CLIPadversarialstickers}.
}, and visualize the attention
of CLIP via Grad-CAM~\cite{2020_ijcv_gradcam}.
It is evident that CLIP pays high attention to the text sticker and accurately understands the meaning of the word, regardless of text variations\footnote{This phenomenon, where CLIP focuses on the text while disregarding the natural object, is also known as \textit{typographic attacks}~\cite{goh2021multimodal}.
Neurons in CLIP image encoders can simultaneously perceive both visual and text signals associated with the same concept, such as an image or typographic text of Spiderman. This ability may stem from the training images containing scene texts in the large training data.
}.
CLIP is trained on massive natural images collected from the web, and its text perception ability may come from the natural images containing scene texts~\cite{goh2021multimodal}.
Will CLIP perceive the text in common STR images~\cite{2013_ic13,2015_karatzas_ic15,zhang2017uber}, which are cropped from a natural image?
Fig.~\ref{fig:str-cam} presents the visualization results of CLIP-ViT-B/32 for STR images.
Although the text in these STR images is occluded, curved, blurred, and rotated, CLIP can still perceive them.
From Fig.~\ref{fig:clip_0shot}\&\ref{fig:str-cam}, we can see CLIP possesses an exceptional capability to perceive and comprehend various text in images.
This is exactly the desired quality for a robust STR backbone.

In this work, we aim to leverage the text perception capability of CLIP for STR and build a strong baseline for future STR research with VLMs.
To this end, we introduce CLIP4STR, a simple yet effective STR framework built upon CLIP.
CLIP4STR consists of two encoder-decoder branches: the visual branch and the cross-modal branch. 
The image and text encoders inherit from CLIP, while the decoders employ the transformer decoder~\cite{2017_attention}. 
To enable the decoder to delve deep into word structures~(dependency relationship among characters in a word), we incorporate the permuted sequence modeling technique proposed by PARSeq~\cite{2022_parseq}. 
This allows the decoder to perform sequence modeling of characters in arbitrary orders without relying on specific sequence order assumptions.
During training, the visual branch provides an initial prediction based on the visual feature, which is then refined by the cross-modal branch to address possible discrepancies between the visual feature and text semantics of the prediction.
The cross-modal branch functions as a semantic-aware spell checker, similar to modern STR methods~\cite{2021_abinet,li2021trocr}. 
For inference, we design a dual predict-and-refine decoding scheme to fully utilize the capabilities of both encoder-decoder branches for improved character recognition.

We scale CLIP4STR across different model sizes, pre-training data, and training data to investigate the effectiveness of large-scale pre-trained VLMs as STR backbones.
CLIP4STR achieves state-of-the-art performance on 13 STR benchmarks, encompassing regular and irregular text.
Additionally, we present a comprehensive empirical study on adapting CLIP to STR.
CLIP4STR provides a simple yet strong baseline for future STR research with VLMs.

\section{Related Work}

\subsection{Vision-Language Models and Its Application} \label{sec:related-vl}
Large-scale pre-trained vision-language models learning under
language supervision such as CLIP~\cite{2021-clip}, ALIGN~\cite{2021-align}, and Florence~\cite{2021_florence} demonstrate excellent generalization abilities.
This encourages researchers to transfer the knowledge of these pre-trained VLMs to different downstream tasks in a fine-tuning or zero-shot fashion.
For instance, \cite{2022_clip4clip,2022_centerclip,wang2022align} tune CLIP on videos and make CLIP specialized in text-video retrieval,
CLIPScore~\cite{2021_CLIPScore} uses CLIP to evaluate the quality of generated image captions,
and \cite{zhao2023rlcf,Cho2022CLIPReward} use CLIP as the reward model during test time or training.
The wide application of VLMs also facilitates the research on different pre-training models, \eg,
ERNIE-ViLG~\cite{2021_ERNIE-ViLG}, CoCa~\cite{2022_coca}, OFA~\cite{2022_ofa},
DeCLIP~\cite{li2021supervision}, FILIP~\cite{2022_filip}, and ALBEF~\cite{2021_li_albef}.
Researchers also explore the power of scaling up the data, \eg, COYO-700M~\cite{kakaobrain2022coyo-700m} and LAION-5B~\cite{2022_laion5b}.
Generally, more data brings more power for large VLMs~\cite{ilharco_gabriel_2021_5143773}.

VLMs pre-trained on large-scale image-text pairs possess many fascinating attributes~\cite{2021-clip,goh2021multimodal,2022_shen_how}.
For instance, some neurons in CLIP can perceive the visual and text signals corresponding to the same concept.
\cite{goh2021multimodal} finds particular neurons in CLIP-RN50$\times$4 respond to both photos of Spiderman and the text \texttt{``spider''} in an image.
This also leads to \textit{Typographic Attacks}, namely, VLMs focus on the text rather than natural objects in an image as shown in Figure~\ref{fig:clip_0shot}.
In this work, we leverage the text perception ability of multi-modal neurons and make CLIP specialize in scene text recognition.

\begin{figure*}[!t]
	\centering
	\includegraphics[width=0.95\linewidth]{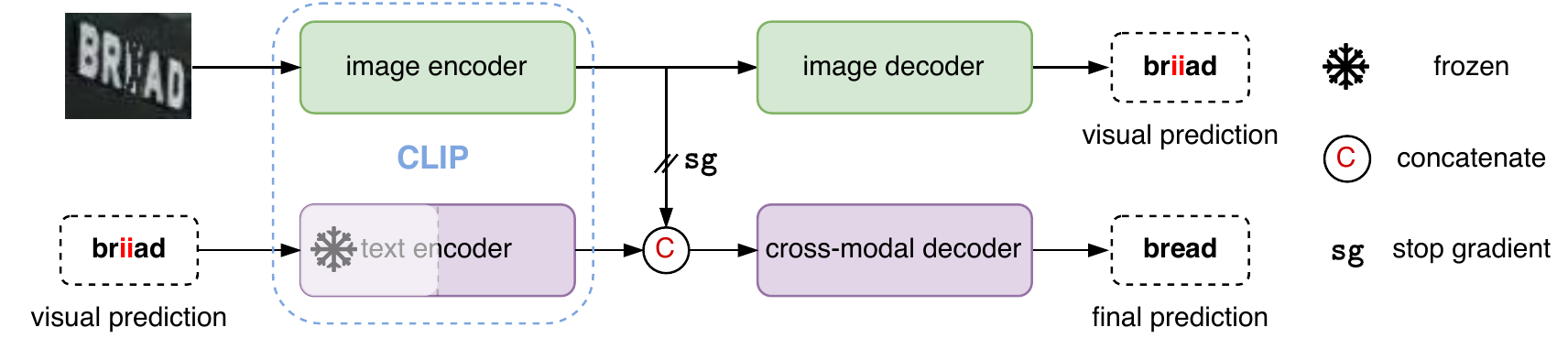}
	\caption{\textbf{The framework of CLIP4STR}. It has a visual branch and a cross-modal branch.
 The cross-modal branch refines the prediction of the visual branch for the final output.
 The text encoder is partially frozen.}
	\label{fig:overall}
\end{figure*}

\subsection{Scene Text Recognition}
Scene text recognition methods can be broadly divided into
two categories: \textit{context-free} and \textit{context-aware}.
Context-free STR methods only utilize the visual features of images, such as CTC-based~\cite{2006_ctc} methods~\cite{2016_he_reading,2017_shi_end2end,2018_fedor_rosetta,atienza2021vision},
segmentation-based methods~\cite{2019_luo_scene,2022_wan_textscanner,2022_zhao_bgi}, and attention-based methods with an encoder-decoder mechanism~\cite{2017_cheng_fan,2019_shi_aster}.
Since context-free STR methods lack the understanding of text semantics, they are less robust against occluded or incomplete text.
Context-aware STR methods are the mainstream approach, leveraging text semantics to enhance recognition performance.
For example, ABINet~\cite{2021_abinet}, LevOCR~\cite{2022_levocr}, MATRN~\cite{2022_matrn}, and TrOCR~\cite{li2021trocr} incorporate an external language model to capture text semantics.
Other methods achieve similar goals with built-in modules, such as RNN~\cite{2016_lee_rrn,gao2021semi}, GRU~\cite{dai2020sloan}, transformer~\cite{2019_NRTR,2022_parseq,fujitake2024dtrocr}.
The context information is interpreted as the relations of textual primitives by Zhang \etal~\cite{zhang2023relational}, who proposes a relational contrastive self-supervised learning STR framework.
Besides the context-free and context-aware methods, some efforts aim to enhance the explainability of STR.
For instance, STRExp~\cite{ty2023scene} utilizes local individual character explanations to deepen the understanding of STR methods.
Moreover, training data plays a vital role in STR.
Traditionally, synthetic data~\cite{2014_max_mj,2016_ankush_sj} are used for training due to the ease of generating a large number of samples.
However, recent research suggests that using realistic training data can lead to better outcomes compared to synthetic data~\cite{2021_baek_what,2022_parseq,jiang2023revisiting,rang2023empirical}.
Motivated by these findings, we primarily employ realistic training data in this work.

The success of VLMs also spreads to the STR area.
For example,  TrOCR~\cite{li2021trocr} adopts separate pre-trained language and vision models plus post-pretraining on STR data in an auto-regressive manner~\cite{radford2019language},
MATRN~\cite{2022_matrn} uses a popular multi-modal fusion manner in VLMs such as ALBEF~\cite{2021_li_albef} and ViLT~\cite{pmlr-v139-kim21k}.
CLIPTER~\cite{aberdam2023clipter} enhances the character recognition performance by utilizing the CLIP features extracted from the global image.
CLIP-OCR~\cite{wang2023symmetrical} leverages both visual and linguistic knowledge from CLIP through feature distillation.
In contrast, we directly transfer CLIP to a robust scene text reader,
eliminating the need for CLIP features from the global image or employing an additional CLIP model as a teacher for the STR reader.
We hope our method can be a strong baseline for future STR research with VLMs.

\section{Method}

\subsection{Preliminary} \label{sec:prelim}
Before illustrating the framework of CLIP4STR,
we first introduce CLIP~\cite{2021-clip} and the permuted
sequence modeling~(PSM) technique proposed by PARSeq~\cite{2022_parseq}.
CLIP serves as the backbone, and
the PSM is used for sequence modeling.

\begin{figure*}[!t]
	\centering
	\includegraphics[width=0.95\linewidth]{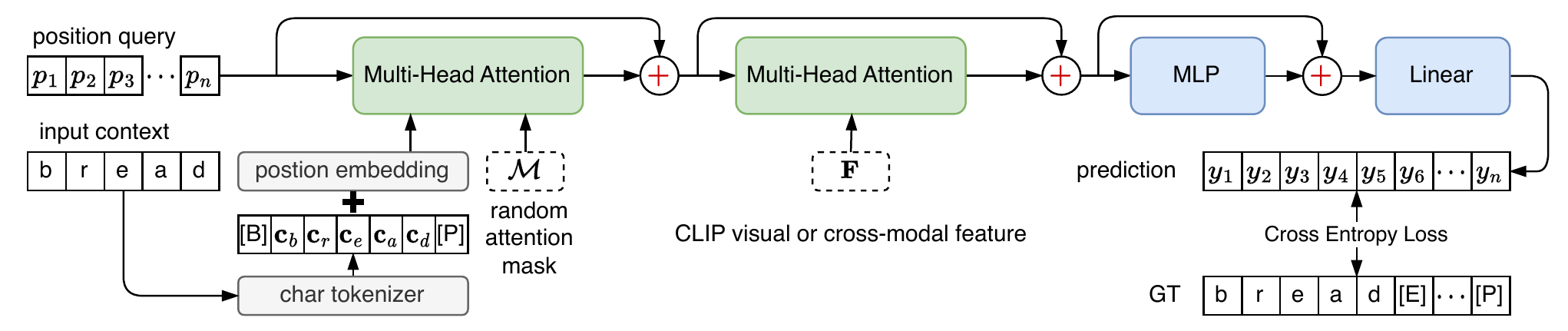}
	\caption{\textbf{The decoder of CLIP4STR}. \texttt{[B]}, \texttt{[E]}, and \texttt{[P]} are the beginning, end, and padding tokens, respectively. `$[\cdots]$' in prediction represents the ignored outputs.
 Layer normalization~\cite{2016_ln}
 and dropout~\cite{2014_dropout} in the decoder are ignored.}
	\label{fig:decoder}
\end{figure*}

\subsubsection{CLIP}
CLIP consists of a text encoder and an image encoder.
CLIP is pre-trained on 400 million image-text pairs using contrastive learning.
The text and image features from CLIP are aligned in a joint image-text embedding space.
\textit{i) The image encoder} of CLIP is a vision transformer (ViT)~\cite{2021_vit}.
Given an image, ViT introduces a visual tokenizer~(convolution) to convert non-overlapped image patches into a discrete sequence.
A [\texttt{CLASS}] token is then prepended to the beginning of the image sequence.
Initially, 
CLIP image encoder only returns the feature of the [\texttt{CLASS}] token,
but in this work, we return features of all tokens.
The rationale behind this choice is that character-level recognition requires fine-grained detail, and local features from all patches are necessary.
These features are normalized
and linearly projected into the joint image-text embedding space.
\textit{ii) The text encoder} of CLIP is a transformer encoder~\cite{2017_attention,2019_bert}.
The text tokenizer is a lower-cased byte pair encoding -- BPE~\cite{2016_bpe} with vocabulary size 49,152.
The beginning and end of the text sequence are padded
with [\texttt{SOS}] and [\texttt{EOS}] tokens, respectively.
Linguistic features of all tokens are utilized for character recognition.
These features are also normalized and linearly projected into the joint image-text embedding space.

\begin{table}[!t]
    \setlength\tabcolsep{0.8pt}
    \caption{\textbf{Examples of attention mask $\mathcal{M}$}.
    The sequences with \texttt{[B]} and \texttt{[E]} represent the input context and output sequence, respectively.
The entry $\mathcal{M}_{i,j}$ = $-\infty$~(negative infinity) indicates that the dependency of output $i$ on input context $j$ is removed.    
    }
    \label{tab:mask}
    \begin{subtable}{0.325\linewidth}
        \centering
        \scalebox{1.0}{
          \begin{tabular}{c|cccc}
            & \texttt{[B]} & $y_1$ & $y_2$ & $y_3$ \\
            \midrule
            $y_1$ & 0 & \scriptsize $-\infty$
            & \scriptsize $-\infty$ & \scriptsize $-\infty$ \\
            $y_2$ & 0 & 0 
            & \scriptsize $-\infty$ & \scriptsize $-\infty$ \\
            $y_3$ & 0 & 0 & 0 & \scriptsize $-\infty$ \\
            \texttt{[E]} & 0 & 0 & 0 & 0 \\
        \end{tabular}}
        \caption{AR mask}
        \label{tab:mask:sub1}
    \end{subtable}
    \begin{subtable}{0.325\linewidth}
        \centering
        \scalebox{1.0}{
          \begin{tabular}{c|cccc}
            & \texttt{[B]} & $y_1$ & $y_2$ & $y_3$ \\
            \midrule
            $y_1$ & 0 & \scriptsize $-\infty$ & 0 & 0 \\
            $y_2$ & 0 & 0 & \scriptsize $-\infty$ & 0 \\
            $y_3$ & 0 & 0 & 0 & \scriptsize $-\infty$ \\
            \texttt{[E]} & 0 & 0 & 0 & 0 \\
        \end{tabular}%
        }
        \caption{cloze mask}
        \label{tab:mask:sub2}
    \end{subtable}
    \begin{subtable}{0.325\linewidth}
        \centering
        \scalebox{1.0}{
          \begin{tabular}{c|cccc}
            & \texttt{[B]} & $y_1$ & $y_2$ & $y_3$ \\
            \midrule
            $y_1$ & 0 & \scriptsize $-\infty$ & 0 & 0 \\
            $y_2$ & 0 & \scriptsize $-\infty$ & \scriptsize $-\infty$ & \scriptsize $-\infty$ \\
            $y_3$ & 0 & \scriptsize $-\infty$ & 0 & \scriptsize $-\infty$ \\
            \texttt{[E]} & 0 & 0 & 0 & 0 \\
        \end{tabular}%
        }
        \caption{random mask}
        \label{tab:mask:sub3}
    \end{subtable}
\end{table}

\subsubsection{Permuted sequence modeling} \label{sec-psm}
Traditionally, STR methods use a left-to-right or right-to-left order to model character sequences \cite{2021_abinet}. 
However, the characters in a word do not strictly follow such directional dependencies.
For instance, to predict the letter ``\texttt{o}'' in the word ``\texttt{model}'', it is sufficient to consider only the context ``\texttt{m\_de}'' rather than relying solely on the left-to-right context ``\texttt{m\_}'' or the right-to-left context ``\texttt{led\_}''. 
The dependencies between characters in a word can take various forms.
To encourage the STR method to explore these structural relationships within words, PARSeq~\cite{2022_parseq} introduces a permuted sequence modeling~(PSM) technique.
This technique uses a random attention mask $\mathcal{M}$ for attention operations~\cite{2017_attention} to generate random dependency relationships between the input context and the output. 
Table~\ref{tab:mask} illustrates three examples of mask $\mathcal{M}$.
We will delve further into this mechanism in \S\ref{sec:dec}.

\subsection{Encoder}
The framework of CLIP4STR is illustrated in Fig.~\ref{fig:overall}.
CLIP4STR employs a dual encoder-decoder design, consisting of a visual branch and a cross-modal branch.
The text and image encoders utilize the architectures and pre-trained weights from CLIP.
The visual branch generates an initial prediction based on the visual features extracted by the image encoder.
Subsequently, the cross-modal branch refines the initial prediction by addressing the discrepancy between the visual features and the textual semantics of the prediction.
Since the image and text features are aligned in a joint image-text embedding space during pre-training, it becomes easy to identify this discrepancy.
The cross-modal branch acts as a semantic-aware spell checker.

The text encoder is partially frozen.
This freezing operation retains the learned text understanding ability of the language model and reduces training costs.
It is a common practice in transfer learning of large language models~\cite{2022_alayrac_flamingo}.
In contrast, the visual branch is fully trainable due to the domain gap between STR data (cropped word images) and CLIP training data (collected from the web, often natural images). 
Additionally, we block the gradient flow from the cross-modal decoder to the visual encoder to enable autonomous learning of the visual branch, resulting in improved refined cross-modal predictions.

For the text encoder $g(\cdot)$ and the image encoder $h(\cdot)$, given the input text $\bm{t}$ and image $\bm{x}$, the text,
image, and cross-modal features are computed as:
\begin{align}
    \bm{F}_t &= g(\bm{t}) \in \mathbb{R}^{L_t\times D}, \\
    \bm{F}_i &= h(\bm{x}) \in \mathbb{R}^{L_i\times D},\\
    \bm{F}_c &= [\bm{F}_i^T ~\bm{F}_t^T]^T \in \mathbb{R}^{L_c\times D},
\end{align}
where $L_t$ represents the text
sequence length, $L_i$ is the sequence length of image tokens,
$D$ denotes the dimension of the joint image-text embedding space,
and the cross-modal sequence length $L_c = L_i + L_t$.

\subsection{Decoder} \label{sec:dec}
The decoder aims to extract the character information
from the visual feature
$\bm{F}_i$ or cross-modal feature $\bm{F}_c$.
The decoder framework is shown in Fig.~\ref{fig:decoder}.
It adopts the design of the transformer
decoder~\cite{2017_attention} plus the PSM technique mentioned in \S~\ref{sec-psm},
enabling a predicted character to have arbitrary dependencies on the input context during training.

The visual and cross-modal decoders have the same architecture but differ in the input.
They receive the following inputs: a learnable position query
$\bm{p} \in \mathbb{R}^{N\times D}$,
an input context $\bm{c} \in \mathbb{R}^{N\times D}$,
and a randomly generated attention mask
$\mathcal{M} \in \mathbb{R}^{N\times N}$.
$N$ represents the length of characters.
The decoder outputs the prediction
$\bm{y} \in \mathbb{R}^{N\times C}$, where $C$ is the number of character classes.
The decoding stage can be denoted as:
    $\bm{y} = \texttt{DEC}(\bm{p}, \bm{c}, \mathcal{M}, \bm{F})$.
The first Multi-Head Attention (MHA) in Fig.~\ref{fig:decoder} performs context-position attention:
\begin{align}
    \bm{m}_1 &= \texttt{softmax}(\frac{\bm{p}\bm{c}^T}{\sqrt{D}} + \mathcal{M})\bm{c} + \bm{p}. \label{eq:mha1}
\end{align}
The second MHA focuses on feature-position attention:
\begin{align}
    \bm{m}_2 = \texttt{softmax}(\frac{\bm{m}_1\bm{F}^T}{\sqrt{D}})\bm{F} + \bm{m}_1. \label{eq:mha2}
\end{align}
For simplicity, we ignore the input and output linear transformations in the attention operations of Eq.~\eqref{eq:mha1}
and Eq.~\eqref{eq:mha2}.
Then $\bm{m}_2 \in \mathbb{R}^{N\times D}$ is used for the final prediction $\bm{y}$:
\begin{align}
    \bm{y} = \texttt{Linear}(\texttt{MLP}(\bm{m}_2) + \bm{m}_2).
\end{align}

During training, the output of the decoder depends on the randomly permuted input context.
This encourages the decoder to analyze the word structure beyond the traditional left-to-right or right-to-left sequence modeling assumptions~\cite{2021_abinet}. 
The inclusion of a random attention mask $\mathcal{M}$ in Eq.\eqref{eq:mha1} enables this capability~\cite{2022_parseq}. 
Table~\ref{tab:mask} presents examples of generated attention masks, including a left-to-right auto-regressive~(AR) mask, a cloze mask, and a random mask.
Following PARSeq~\cite{2022_parseq}, we employ $K=6$ masks per input context during training.
The first two masks are left-to-right and right-to-left masks, and others are randomly generated.
CLIP4STR is optimized to minimize the sum of cross-entropy losses ($\texttt{CE}(\cdot)$) of the visual branch and the cross-modal branch:
\begin{align}
    \mathcal{L} = \texttt{CE}(\bm{y}^i, \hat{\bm{y}}) + \texttt{CE}(\bm{y}, \hat{\bm{y}}),
\end{align}
where $\hat{\bm{y}}$, $\bm{y}^i$, and $\bm{y}$ indicate ground truth, prediction of the visual branch,
and prediction of the cross-modal branch.

\begin{algorithm}[!t]\small
	\caption{Inference decoding scheme (\S\ref{app:inference})}
	\label{algo-decode}
	\KwIn{image $\bm{x}$,
    image encoder $h(\cdot)$ and decoder $\texttt{Dec}^i(\cdot)$,
    text encoder $g(\cdot)$, cross-modal decoder $\texttt{Dec}^c(\cdot)$,
    AR mask $\mathcal{M}^a$, cloze mask $\mathcal{M}^c$,
    image and cross-modal position query $\bm{p}^i$ and $\bm{p}^c$,
    context $\bm{c} = \bm{0} \in \mathbb{R}^{N\times D}$,
    char and text tokenizer $\texttt{CTK}(\cdot)$ and $\texttt{TTK}(\cdot)$,
    iterative refinement times $T_i$}
	\KwOut{prediction $\bm{y}$}

        \tcp{$\bm{c}_{1,\cdot}$ denote the $1$st row}
        $\bm{c}_{1,\cdot} \leftarrow \texttt{CTK}(\texttt{[B]})$\;
        $\bm{F}_i \leftarrow h(\bm{x})$\;
        \tcp{autoregressive visual decode}
        $\bm{y}^i \leftarrow \bm{0}$\;
        \For{$k\leftarrow 1$ \KwTo $N - 1$}
        {
            $\bm{y}^i_{k,\cdot} \leftarrow \texttt{Dec}^i(\bm{p}^i_{k, \cdot}, \bm{c}_{1:k,\cdot}, \mathcal{M}^a_{1:k,1:k}, \bm{F}_i) $\;
            $\bm{c}_{k+1,\cdot} \leftarrow \texttt{CTK}(\bm{y}^i_{k, \cdot})$\;
        }
        \tcp{autoregressive cross-modal decode}
        $\bm{F}_c \leftarrow [\bm{F}_i^T ~g(\texttt{TTK}(\bm{y}^i))^T]^T$\;
        $\bm{y} \leftarrow \bm{0}$\;
        \For{$k\leftarrow 1$ \KwTo $N - 1$}
        {
            $\bm{y}_{k,\cdot} \leftarrow \texttt{Dec}^c(\bm{p}^c_{k, \cdot}, \bm{c}_{1:k,\cdot}, \mathcal{M}^a_{1:k,1:k}, \bm{F}_c) $\;
            $\bm{c}_{k+1,\cdot} \leftarrow \texttt{CTK}(\bm{y}_{k, \cdot})$\;
        }
        \tcp{refinement with cloze mask}
        \For{$k\leftarrow 1$ \KwTo $T_i$}
        {
            $\bm{c} \leftarrow [\texttt{CTK}(\texttt{[B]})^T ~\texttt{CTK}(\bm{y}_{1:N-1, \cdot}^i)^T]^T$\;
            $\bm{y}^i \leftarrow \texttt{Dec}^i(\bm{p}^i, \bm{c}, \mathcal{M}^c, \bm{F}_i) $\;
            $\bm{F}_c \leftarrow [\bm{F}_i^T ~g(\texttt{TTK}(\bm{y}^i))^T]^T$\;
            $\bm{c} \leftarrow [\texttt{CTK}(\texttt{[B]})^T ~\texttt{CTK}(\bm{y}_{1:N-1, \cdot})^T]^T$\;
            $\bm{y} \leftarrow \texttt{Dec}^c(\bm{p}^c, \bm{c}, \mathcal{M}^c, \bm{F}_c) $\;
        }
\end{algorithm}

\subsubsection{Decoding scheme}
CLIP4STR consists of two branches: a visual branch and a cross-modal branch.
To fully exploit the capacity of both branches, we design a \textit{dual predict-and-refine} decoding scheme for inference, inspired by previous STR methods~\cite{2021_abinet,2022_parseq}. 
Alg.~\ref{algo-decode} illustrates the decoding process.
The visual branch first performs autoregressive decoding, where the future output depends on previous predictions.
Subsequently, the cross-modal branch addresses possible discrepancies between the visual feature and the text semantics of the visual prediction, aiming to improve recognition accuracy.
This process is also autoregressive.
Finally, the previous predictions are utilized as the input context for refining the output in a cloze-filling manner.
The refinement process can be iterative.
After iterative refinement, the output of the cross-modal branch serves as the final prediction.


\section{Experiment}

\begin{table}[!t]
  \caption{\textbf{Model sizes and optimization hyper-parameter}.
The learning rate for CLIP encoders is $\text{8.4e-5}\times\frac{\text{batch}}{512}$~\cite{2017_goyal_accurate}.
For models trained from scratch~(decoders), the learning rate is multiplied by 19.0.
Params is the total parameters in a model, and non-trainable parameters in three models are 44.3M, 80.5M, and
126M, respectively.
Training time is measured on 8 NVIDIA RTX A6000 GPUs.
  }
  \label{tab:exp-detail}
  \centering
  \resizebox{0.98\linewidth}{!}{%
  \begin{tabular}{l|rlcccc}
    \toprule
    {Model} & {Params} & Train Data & {Batch} & {Epochs} & Time\\
    \midrule
    CLIP4STR-B & 158M & Real(3.3M) & 1024  & 16 & 12.8h \\
    CLIP4STR-L & 446M & Real(3.3M) & 1024  & 10 & 23.4h \\
    CLIP4STR-H & 1B & RBU(6.5M) & 1024 & 4 & 48.0h \\
    \bottomrule
  \end{tabular}
}
\end{table}

  

\subsection{Experimental Details} \label{sec:exp-details}
\noindent
We instantiate CLIP4STR with CLIP-ViT-B/16, CLIP-ViT-L/14, and CLIP-ViT-H/14~\cite{ilharco_gabriel_2021_5143773}.
Table~\ref{tab:exp-detail} presents the main hyper-parameters of CLIP4STR.
A reproduction of CLIP4STR is at {\color{blue}\texttt{{
\url{https://github.com/VamosC/CLIP4STR}.
}}
}

\noindent
\textbf{Test benchmarks~}
The evaluation benchmarks include IIIT5K~\cite{2012_IIT5k},
CUTE80~\cite{2014_cute80},
Street View Text~(SVT)~\cite{2011_wang_svt},
SVT-Perspective~(SVTP)~\cite{2013_phan_svtp},
ICDAR 2013~(IC13)~\cite{2013_ic13},
ICDAR 2015~(IC15)~\cite{2015_karatzas_ic15},
and three occluded datasets -- HOST, WOST~\cite{2021_wang_ost}, and OCTT~\cite{raisi2022occluded}.
Additionally, we utilize 3 recent large 
benchmarks: COCO-Text~(low-resolution, occluded text)~\cite{2016_andreas_cocotext},
ArT~(curved and
rotated text)~\cite{2019_icdar_art},
and Uber-Text~(vertical and rotated text)~\cite{zhang2017uber}.

\noindent
\textbf{Training dataset~}
{\textbf{1)}} \textbf{MJ+SJ}:
MJSynth (MJ, 9M samples)~\cite{2014_max_mj} and SynthText (ST, 6.9M samples)~\cite{2016_ankush_sj}.
{\textbf{2)}} \textbf{Real(3.3M)}:
COCO-Text~(COCO)~\cite{2016_andreas_cocotext},
RCTW17~\cite{2017_shi_rctw},
Uber-Text~(Uber)~\cite{zhang2017uber},
ArT~\cite{2019_icdar_art},
LSVT~\cite{2019_sun_lsvt},
MLT19~\cite{2019_nayef_mlt},
ReCTS~\cite{2019_zhang_rects},
TextOCR~\cite{2021_singh_textocr},
Open Images~\cite{openimages} annotations from the OpenVINO toolkit~\cite{2021_krylov_openimagev5}.
These real datasets have 3.3M images in total.
{\textbf{3)}} \textbf{RBU(6.5M)}:
A dataset provided by~\cite{rang2023empirical}.
It combines the Real(3.3M), benchmark datsets~(training data of SVT, IIIT5K, IC13, and IC15), and part of Union14M-L~\cite{jiang2023revisiting}.

\noindent
\textbf{Learning strategies~}
We apply a warm up and
cosine learning rate decay policy.
The batch size is kept to be close to 1024.
For large models, this is achieved by gradient accumulation.
For synthetic data, we train CLIP4STR-B for 6 epochs and CLIP4STR-L for 5 epochs.
For RBU(6.5M) data, we train 11, 5, and 4 epochs for CLIP4STR-B, CLIP4STR-L, and CLIP4STR-H, respectively.
AdamW~\cite{2019_ilya_adamw} optimizer is adopted with
a weight decay value 0.2.
All experiments are performed with mixed precision~\cite{2018_iclr_amp}.

\begin{table*}[!t]
    \centering
    \caption{\textbf{Word accuracy on 10 common benchmarks.
    The\colorbox{Gray1}{\textbf{best}}and\colorbox{Gray2}{second-best}results are highlighted.
    Benchmark datasets~(\textbf{B}) - SVT, IIIT5K, IC13, and IC15.
    `N/A' for not applicable.
    $^\sharp$Reproduced by PARSeq~\cite{2022_parseq}.
    }}
    \setlength\tabcolsep{4pt}
    \scalebox{1.05}{
    \begin{tabular}{l|l|l|cccccccccccc}
    \toprule
    \multirow{2}{*}{Method} 
    & \multirow{2}{*}{Pre-train Data}  & \multirow{2}{*}{Train Data}
    & IIIT5K & SVT & IC13 & IC15 & IC15 & SVTP & CUTE
    & HOST & WOST & OCTT \\
    &  &  & 3,000 & 647 & 1,015 
    & 1,811 & 2,077 & 645 & 288 &2,416
    & 2,416 & 1,911\\
    \midrule
    
    0-shot CLIP~\cite{2021-clip} & WIT 400M~\cite{2021-clip} & N/A &
    90.0 & -- & -- & -- & -- & -- & -- & -- & -- & --\\
    \midrule

    SRN~\cite{2020_yu_SRN} & ImageNet-1K & MJ+ST
    & 94.8 & 91.5 & -- & 82.7 & -- & 85.1 & 87.8 & -- & -- & --\\
    
    TextScanner~\cite{2022_wan_textscanner} & N/A & MJ+ST
    & 95.7 & 92.7 & 94.9 & -- & 83.5 & 84.8 & 91.6 & -- & -- & --\\
    

    RCEED~\cite{2021_cui_RCEED} & N/A & MJ+ST+B
    & 94.9 & 91.8 & -- & -- & 82.2 & 83.6 & 91.7 & -- & -- & --\\

    TRBA~\cite{2021_baek_what} & N/A & MJ+ST 
    & 92.1 & 88.9 & -- & 86.0 & -- & 89.3 & 89.2 & -- & -- & --\\

    VisionLAN~\cite{2021_wang_ost} & From Scratch & MJ+ST 
    & 95.8 & 91.7 & -- & 83.7 & -- & 86.0 & 88.5 & 50.3 & 70.3 & --\\
    

    ABINet~\cite{2021_abinet} & WikiText-103 & MJ+ST
    & 96.2 & 93.5 & -- & 86.0 & -- & 89.3 & 89.2 & -- & -- & --\\

    ViTSTR-B~\cite{atienza2021vision} & ImageNet-1K & MJ+ST
    & 88.4 & 87.7 & 92.4 & 78.5 & 72.6 & 81.8 & 81.3 & -- & -- & --\\

    LevOCR~\cite{2022_levocr} & WikiText-103 & MJ+ST 
    & 96.6 & 92.9 & -- & 86.4 & -- & 88.1 & 91.7 & -- & -- & --\\

    MATRN~\cite{2022_matrn} & WikiText-103 & MJ+ST
    & 96.6 & 95.0 & 95.8 & 86.6 & 82.8 & 90.6 & 93.5 & -- & -- & --\\

    PETR~\cite{wang2022petr} & N/A & MJ+ST
    & 95.8 & 92.4 
    & 97.0 & 83.3 & -- & 86.2 & 89.9 & -- & -- & --\\

    DiG-ViT-B~\cite{yang2022reading} & Textimages-33M & MJ+ST 
    & 96.7 & 94.6 & 96.9 & 87.1 & -- & 91.0 & 91.3 & 74.9 & 82.3 & --\\

    PARSeq$_A$~\cite{2022_parseq} & From Scratch & MJ+ST
    & 97.0 & 93.6 & 96.2 & 86.5 & 82.9 & 88.9 & 92.2 & -- & -- & --\\

    TrOCR$_{Large}$~\cite{li2021trocr} & Textlines-684M & MJ+ST+B
    & 94.1 & 96.1 
    & 97.3 & 88.1 & 84.1 & 93.0 & 95.1 & -- & -- & --\\

    SIGA$_{T}$~\cite{guan2023self} & ImageNet-1K & MJ+ST
    & 96.6 & 95.1 & 96.8 & 86.6 & 83.0 & 90.5 & 93.1 & -- & -- & --\\
    CLIP-OCR~\cite{wang2023symmetrical} & From Scratch & MJ+ST
    & 97.3 & 94.7 & -- & 87.2 & -- & 89.9 & 93.1 & -- & -- & -- \\
    LISTER-B~\cite{cheng2023lister} & N/A & MJ+ST
    & 96.9 & 93.8 & -- & 87.2 & -- & 87.5 & 93.1 & -- & -- & -- \\
    CLIPTER~\cite{aberdam2023clipter} & N/A & Real(1.5M)
    & -- & 96.6 & -- & -- & 85.9 & -- & -- & -- & -- & -- \\
    
    DiG-ViT-B~\cite{yang2022reading} & Textimages-33M & Real(2.8M) 
    & 97.6 & 96.5 & 97.6 & 88.9 & -- & 92.9 & 96.5 & 62.8 & 79.7 & -- \\
    CCD-ViT-B~\cite{guan2023selfccd} & Textimages-33M & Real(2.8M) 
    & 98.0 & 97.8 & 98.3 & 91.6 & -- & 96.1 & 98.3 & -- & -- & -- \\

    ViTSTR-S~\cite{atienza2021vision}$^\sharp$ & ImageNet-1K & Real(3.3M)
    & 97.9 & 96.0 & 97.8 & 89.0 & 87.5 & 91.5 & 96.2 & 64.5 & 77.9 & 64.2\\

    ABINet~\cite{2021_abinet}$^\sharp$ & From Scratch & Real(3.3M)
    & 98.6 & 98.2 & 98.0 & 90.5 & 88.7 & 94.1 & 97.2 & 72.2 & 85.0 & 70.1\\
    
    PARSeq$_A$~\cite{2022_parseq} & From Scratch & Real(3.3M)
    & 99.1 & 97.9 & 98.4
    & 90.7 & 89.6 & 95.7 & 98.3 & 74.4 & 85.4 & 73.1 \\

    MAERec-B~\cite{jiang2023revisiting} & Union14M-U & Union14M-L
    & 98.5 & 97.8 
    & 98.1
    & -- & 89.5 & 94.4 & 98.6 & -- & -- & -- \\

    \midrule
    CLIP4STR-B & \multirow{4}{*}{WIT 400M}
    & MJ+ST
    & 97.7 & 95.2 & 96.1 & 87.6 & 84.2 & 91.3 & 95.5
    & 79.8 & 87.0 & 57.1 \\

    CLIP4STR-L & & MJ+ST
    & 98.0 & 95.2 & 96.9 & 87.7 & 84.5 & 93.3 & 95.1 
    & \cellcolor{Gray1}\textbf{82.7}
    & 88.8 & 59.2 \\

    CLIP4STR-B & & Real(3.3M)
    & 99.2 
    & 98.3 
    & 98.3 
    & {91.4}
    & {90.6} 
    & {97.2} 
    & \cellcolor{Gray2}{99.3}
    & 77.5 
    & 87.5 & 81.8 \\

   CLIP4STR-L & & Real(3.3M)
    & \cellcolor{Gray2}{99.5}
    & {98.5}
    & {98.5}
    & 91.3 
    & {90.8} 
    & {97.4} 
    & 99.0
    & 79.8 
    & {89.2} 
    & 84.9\\

    \hdashline[1pt/1pt]
    CLIP4STR-B &\multirow{3}{*}{DataComp-1B~\cite{gadre2024datacomp}}  & Real(3.3M)
    & 99.4
    & \cellcolor{Gray2}98.6
    & 98.3 
    & 90.8
    & 90.3 
    & 97.8
    & 99.0
    & 77.6 
    & 87.9 & 83.1 \\

    CLIP4STR-B & & RBU(6.5M)
    & \cellcolor{Gray2}99.5
    & 98.3
    & 98.6
    & 91.4
    & 91.1
    & 98.0
    & 99.0
    & 79.3 
    & 88.8 
    & 83.5\\

    CLIP4STR-L & & RBU(6.5M)
    & \cellcolor{Gray1}\textbf{99.6}
    & \cellcolor{Gray2}98.6
    & \cellcolor{Gray1}\textbf{99.0}
    & \cellcolor{Gray1}\textbf{91.9}
    & \cellcolor{Gray1}\textbf{91.4}
    & \cellcolor{Gray1}\textbf{98.1}
    & \cellcolor{Gray1}\textbf{99.7}
    & 81.1 
    & \cellcolor{Gray2}90.6 
    & \cellcolor{Gray2}85.9\\

    CLIP4STR-H & DFN-5B~\cite{fang2023data} & RBU(6.5M)
    & \cellcolor{Gray2}99.5
    & \cellcolor{Gray1}\textbf{99.1}
    & \cellcolor{Gray2}98.9
    & \cellcolor{Gray2}91.7
    & \cellcolor{Gray2}91.0
    & \cellcolor{Gray2}98.0
    & 99.0
    & \cellcolor{Gray2}82.6 
    & \cellcolor{Gray1}\textbf{90.9} 
    & \cellcolor{Gray1}\textbf{86.5} \\

    \bottomrule
  \end{tabular}}
  \label{tab:main-results}
  \vspace{-1em}
\end{table*}

\noindent
\textbf{Data and label processing~}
RandAugment~\cite{2020_ekin_randaug} excludes sharpness and invert is used
with layer depth 3 and magnitude 5.
The image size is 224$\times$224.
The sequence length of the text encoder is 16.
The maximum length of the character sequence is 25.
Considering an extra \texttt{[B]} or \texttt{[E]} token, we set $N=26$.
During training, the number of character classes $C=94$, \ie,
mixed-case alphanumeric characters and punctuation marks are recognized.
During inference, we only use a lowercase alphanumeric charset, \ie, $C=36$. The iterative refinement times $T_i=1$.
The evaluation metric is word accuracy.

\subsection{Comparison to State-of-the-art}
We compare CLIP4STR with previous SOTA methods on 10 common STR benchmarks in Table~\ref{tab:main-results}.
CLIP4STR surpasses the previous methods by a significant margin, achieving new SOTA performance.
Notably, CLIP4STR performs exceptionally well on irregular text datasets, such as IC15~(incidental scene text), SVTP~(perspective scene text), CUTE~(curved text line images), HOST~(heavily occluded scene text), and WOST~(weakly occluded scene text). 
This aligns with the examples shown in Fig.~\ref{fig:clip_0shot}\&\ref{fig:str-cam} and supports our motivation for adapting CLIP as a scene text reader, as CLIP demonstrates robust identification of regular and irregular text.
CLIP4STR exhibits excellent reading ability on occluded datasets, surpassing the previous SOTA by \textbf{7.8\%} and \textbf{5.4\%} in the best case on HOST and WOST, respectively.
This ability can be attributed to the pre-trained text encoder and cross-modal decoder, which can infer missing characters using text semantics or visual features.
The performance of CLIP4STR is also much better than CLIP-OCR~\cite{wang2023symmetrical} and CLIPTER~\cite{aberdam2023clipter}, both of which work in a similar direction as CLIP4STR.
This demonstrates that directly transferring CLIP into a STR reader is more effective than the distillation method~\cite{wang2023symmetrical} or utilizing CLIP features of the global image as auxiliary context~\cite{aberdam2023clipter}.

\begin{table}[!t]
    \centering
    \caption{
    \textbf{Word accuracy on 3 large benchmarks}.\\
    $^\sharp$Reproduced by PARSeq~\cite{2022_parseq}.
    }
    \label{tab:3large}
    \scalebox{1.08}{
    \begin{tabular}{l|c|ccc}
    \toprule
    \multirow{2}{*}{Method} & Train
    & COCO & ArT & Uber   \\
    & data & 9,825 & 35,149 & 80,551 \\
    \midrule


    ViTSTR-S~\cite{atienza2021vision}$^\sharp$ & MJ+ST
    & 56.4 & 66.1 & 37.6 \\
    
    TRBA~\cite{2021_baek_what}$^\sharp$ & MJ+ST
    & 61.4 & 68.2 & 38.0 \\
    
    ABINet~\cite{2021_abinet}$^\sharp$ & MJ+ST
    & 57.1 & 65.4 & 34.9 \\
    
    PARSeq$_A$~\cite{2022_parseq} & MJ+ST
    & 64.0 & 70.7 & 42.0  \\

    MPSTR$_A$~\cite{yang2023masked} & MJ+ST
    & 64.5 & 69.9  & 42.8  \\
    CLIP-OCR~\cite{wang2023symmetrical} & MJ+ST
    & \cellcolor{Gray2}66.5 & 70.5 & 42.4 \\
    \hdashline[1pt/1pt]

    CLIP4STR-B & MJ+ST
    & 66.3 
    & \cellcolor{Gray2}72.8  
    & \cellcolor{Gray2}43.4 \\

    CLIP4STR-L & MJ+ST
    & \cellcolor{Gray1}\textbf{67.0}
    & \cellcolor{Gray1}\textbf{73.7}  
    & \cellcolor{Gray1}\textbf{44.5} \\

    \midrule
    DiG-ViT-B~\cite{yang2022reading} & Real(2.8M) 
    & 75.8 & -- & --  \\

    ViTSTR-S~\cite{atienza2021vision}$^\sharp$ & Real(3.3M)
    & 73.6 & 81.0 & 78.2 \\

    TRBA~\cite{2021_baek_what}$^\sharp$ & Real(3.3M)
    & 77.5 & 82.5 & 81.2 \\

    ABINet~\cite{2021_abinet}$^\sharp$ & Real(3.3M)
    & 76.5 & 81.2  & 71.2 \\
    
    PARSeq$_A$~\cite{2022_parseq} & Real(3.3M)
    & 79.8 & 84.5  & 84.1  \\

    MPSTR$_A$~\cite{yang2023masked} & Real(3.3M)
    & 80.3 & 84.4  & 84.9  \\
    \hdashline[1pt/1pt]

    CLIP4STR-B & Real(3.3M)
    & {81.1}
    & {85.8} 
    & {86.8} \\

    CLIP4STR-L & Real(3.3M)
    & 81.9
    & \cellcolor{Gray2}85.9
    & 87.6 \\
    
    CLIP4STR-B & RBU(6.5M)
    & 81.3
    & 85.8 
    & \cellcolor{Gray2}92.1 \\ 

    CLIP4STR-L & RBU(6.5M)
    & \cellcolor{Gray1}82.7
    & \cellcolor{Gray1}86.4 
    & \cellcolor{Gray1}\textbf{92.2} \\ 

        CLIP4STR-H & RBU(6.5M)
    & \cellcolor{Gray1}\textbf{83.0}
    & \cellcolor{Gray1}\textbf{86.4} 
    & 91.7 \\
    \bottomrule
  \end{tabular}}
  \vspace{-1em}
\end{table}

In addition to the small-scale common benchmarks, we also evaluate CLIP4STR on 3 larger and more challenging benchmarks.
These benchmarks primarily consist of irregular texts with various shapes, low-resolution images, rotation, \etc.
The results, shown in Table~\ref{tab:3large}, further demonstrate the strong generalization ability of CLIP4STR.
CLIP4STR substantially outperforms the previous SOTA methods on these three large and challenging benchmarks.
At the same time, we observe that scaling CLIP4STR to 1B parameters does not bring much improvement in performance.
CLIP4STR-L is comparable to CLIP4STR-H in most cases, while
CLIP4STR-H is superior in recognizing occluded characters~(WOST, HOST, OCTT).

\section{Empirical Study}
This section presents our empirical study on adapting CLIP to STR.
Without mention, the models are all trained on 3.3M real data, and the IC15 dataset here contains 2,077 samples.
The average accuracy reported in this section is calculated over the first 9 benchmarks~(14,315 samples) in Table~\ref{tab:main-results}.

\begin{table}[!t]
    \centering
    \caption{\textbf{Ablation study of different components of CLIP4STR}.
    PSM is short for the permuted sequence modeling technique~\cite{2022_parseq}.
    Recipe represents the training recipe for CLIP4STR in \S\ref{sec:exp-details}.
    Cross denotes the cross-modal branch.
    \texttt{[CLASS]} with a \cmark mark means the decoders only use the \texttt{[CLASS]} and \texttt{[EOS]} of CLIP encoders rather than features of all tokens~(refer to \S\ref{sec:prelim}).
    }
    \setlength\tabcolsep{4pt}
    \scalebox{1.0}{
    \begin{tabular}{ccccccc|c}
    \toprule
    \multicolumn{7}{l}{Reference Method}  & Avg. \\
    \midrule
    \multicolumn{7}{l}{ABINet~\cite{2021_abinet}} & 89.1 \\
    \multicolumn{7}{l}{PARSeq$_A$~\cite{2022_parseq}} (previous SOTA) & 89.9 \\
    \midrule
    Base & PSM & ViT-B & Recipe & Cross & \texttt{[CLASS]}& ViT-L & Avg.   \\
    \midrule
    \cmark & & & & & & & 89.2 \\
    \cmark & \cmark & & & & & & 89.9 \\
    \cmark & \cmark & \cmark & & & & & 90.0 \\
    \cmark & \cmark & \cmark & \cmark & & & & 90.8 \\
    \cmark & \xmark & \cmark & \cmark & & & & 90.0\\
    \cmark & \cmark & \cmark & \cmark & \cmark & \cmark & & 90.6 \\
    \cmark & \cmark & \cmark & \cmark & \cmark & \xmark & & 91.2 \\
    \cmark & \cmark & \cmark & \cmark & \cmark & \xmark & \cmark
    & \cellcolor{Gray1}{91.9} \\    
    \bottomrule
  \end{tabular}}
  \label{tab:abla}
\end{table}

\subsection{Ablation Study of CLIP4STR}
Table~\ref{tab:main-results}\&\ref{tab:3large} show that
CLIP4STR achieves SOTA performance on 11 STR benchmarks.
What are the sources of this high performance?
We conduct ablation studies of different components in Table~\ref{tab:abla}, starting with the visual branch in Fig.~\ref{fig:overall} as the baseline~(accuracy 89.2\%).
The encoder is a ViT-S without pre-training.
Then we apply the permuted sequence modeling~(PSM) technique~\cite{2022_parseq} to the visual decoder and follow the training recipe of PARSeq:
4$\times$8 patch size, the same learning rate for the encoder and decoder, and 20 training epochs.
This brings a 0.7\% improvement in accuracy.
Next, we replace the encoder with the image encoder of CLIP-ViT-B/16.
However, no significant gain is observed without adaptations.
To unleash the potential of CLIP, we adjust the training recipe: using 16$\times$16 patch size, a small learning rate for CLIP encoders, a relatively large learning rate for decoders, and fewer training epochs --- 16~(\S\ref{sec:exp-details}).
The learning rate is searched automatically by Ray~\cite{moritz2018ray}, and the best number of training epochs is decided by manual test.
CLIP makes the model converge easier and faster, so the training recipe should change accordingly.
To better grasp the contribution of PSM, we conducted an experiment where we removed PSM and achieved an accuracy of 90.0\%.

Although the performance of the visual branch is already very high~(90.8\%), the cross-modal branch further improves the accuracy by 0.4\%, demonstrating its effectiveness.
It is worth noting that utilizing CLIP features of all patches is crucial for character recognition.
Only using the \texttt{[CLASS]} and \texttt{[EOS]} results in inferior performance -- 90.6\%.
Moreover, the use of a large model --- CLIP-ViT-L/14 further increases the accuracy by 0.7\%.
The large CLIP-ViT-L/14 converges faster than CLIP-ViT-B/16 for STR. It only requires 10 epochs of training on the Real(3.3M) data, much less than the training epochs of CLIP-ViT-B/16.

\begin{table}[!t]
    \centering
    \caption{\textbf{Freezing options in CLIP4STR-B}.
    \#Params means the number of \textit{learnable}
    parameters of encoders in CLIP4STR-B.
    One decoder in CLIP4STR-B has 4.3M parameters.
    \texttt{token} means we only use pre-trained token embeddings of CLIP text encoder
    as text features.
    }
    \setlength\tabcolsep{4pt}
    \scalebox{1.0}{
    \begin{tabular}{ccc|ccccc}
    \toprule
    \multicolumn{2}{c}{Frozen Layers}
    &  \multirow{2}{*}{\centering \#Params}
    &  \multirow{2}{*}{\centering IC15}
    &  \multirow{2}{*}{\centering WOST}
    &  \multirow{2}{*}{\centering HOST}
    &  \multirow{2}{*}{\centering COCO}
    &  \multirow{2}{*}{\centering Uber} \\
    Image & Text & & & & & \\
    \midrule
    0 & 0 & 149~M 
    & \cellcolor{Gray1}{90.8} & 87.5 & 76.4 & 80.8 
    & \cellcolor{Gray1}{87.0} \\
    0 & 3 & 114~M 
    &  90.4 & \cellcolor{Gray1}{88.1} & 76.9 
    & \cellcolor{Gray1}{81.2} & 86.8 \\
    0 & 6 & 104~M 
    & 90.6  & 87.5 & \cellcolor{Gray1}{77.5} & 81.1 & 86.8 \\
    0 & 9 & 95~M 
    & 90.3  & 86.8 & 74.9 & 80.9 & 86.3 \\
    0 & 12 & 86~M 
    & 90.3 & 86.1 & 74.9 & 80.9 & 86.4\\

    0 & \texttt{token} & 86~M
    & 90.7 & 87.3 & 77.0 & 80.9 & 86.7 \\
    
    \midrule
    0 & 6 & 95~M 
    & \cellcolor{Gray1}{90.6} & 87.5 & \cellcolor{Gray1}{77.5} & 81.1 & \cellcolor{Gray1}{86.8} \\
    3 & 6 & 84~M 
    & 90.4  & \cellcolor{Gray1}{88.5} & 76.5 & \cellcolor{Gray1}{81.3} & 86.4 \\
    6 & 6 & 62~M 
    & 89.5  & 86.7 & 72.8 & 80.3 & 83.8 \\
    9 & 6 & 41~M 
    & 87.8  & 80.0 & 64.0 & 75.3 & 72.8 \\
    12 & 6 & 19~M 
    & 61.2 & 55.8 & 40.4 & 49.5 & 20.6 \\
    \bottomrule
  \end{tabular}}
  \label{tab:freeze}
\end{table}

\subsection{Parameter Freezing Options}

In CLIP4STR, we freeze half of the layers in the CLIP text encoder, which is a common practice when transferring a large language model to new tasks~\cite{2022_alayrac_flamingo}.
Table~\ref{tab:freeze} illustrates the influence of different parameter freezing options.
The results indicate that freezing the language model has a lesser impact compared to freezing the image model.
Despite using the fixed pre-trained token embeddings of the CLIP text encoder, the system can still achieve satisfactory performance.
This demonstrates that semantic understanding in STR is relatively easier compared to general language understanding.
In STR, text mainly consists of words and phrases, which simplifies the task compared to the general language case.
On the other hand, freezing the image models has a significant impact on performance. The substantial domain gap between the data in STR and the pre-trained data of the CLIP image encoder possibly contributes to this discrepancy.
CLIP is pre-trained on web images, which are primarily natural images.
In contrast, the scene text recognition data comprises cropped word images.
Such a disparity may necessitate a fully trainable image encoder in CLIP4STR to bridge the domain gap.

\subsection{Comparison to Single-modality Pre-trained Model}

In previous empirical studies, we see the effectiveness of CLIP as a STR backbone.
Is VLM better than models pre-trained on single-modality data?
To further clarify this question, Table~\ref{tab:pretrained} presents the results of replacing the visual encoder in Fig.~\ref{fig:overall} with a random initialized ViT,
an ImageNet-1K~\cite{2015_loga_imagenet} pre-trained ViT via DeiT~\cite{2021_hugo_deit}\footnote{\url{https://github.com/facebookresearch/deit}},
and an ImageNet-21K pre-trained ViT
provided by Ridnik \textit{et al.}~\cite{ridnik2021imagenet}.
The training schedules including the learning rate and training epochs are kept the same as CLIP4STR.
In Table~\ref{tab:pretrained}, the ImageNet pre-trained models even perform worse than the model trained from scratch.
Previous works also support this finding.
PARSeq~\cite{2022_parseq} trains its vision transformer from scratch rather than using a pre-trained model.
TrOCR~\cite{li2021trocr} uses pre-trained transformers from DeiT~\cite{2021_hugo_deit}, BEiT~\cite{2022_bao_beit}, and RoBERTa~\cite{2019_roberta}, but it still post-pretrains them on 684M textlines from publicly available PDF files on the Internet.

ImageNet classification pre-training does not align well with STR.
Classifying objects in an image does not help the model learn specific information about the text within the image. For example, two images -- one of a cat and one of a dog -- both containing the text ``\texttt{park}'' cause the model to learn contradictory information about the same text.
In contrast, the vision encoders in CLIP can accurately perceive text signals due to the presence of multi-modal neurons~\cite{goh2021multimodal}~(\S\ref{sec:related-vl}), making CLIP a strong backbone for STR.

\begin{table}[!t]
    \centering
    \caption{\textbf{Different pre-training strategies}.
    \#Params means the learnable parameters in the visual encoder.
    For a fair comparison, only the results of the visual branch in CLIP4STR-B are shown.
    }
    \setlength\tabcolsep{3pt}
    \scalebox{1.08}{
    \begin{tabular}{lc|cccccc}
    \toprule
    Pre-train
    &  \multirow{1}{*}{\centering \#Params}
    &  \multirow{1}{*}{\centering IC15}
    &  \multirow{1}{*}{\centering WOST}
    &  \multirow{1}{*}{\centering HOST}
    &  \multirow{1}{*}{\centering COCO} 
    & \multirow{1}{*}{\centering Uber} \\
    \midrule
     Scratch & 86~M & 90.1 & 84.9 & 74.8 & 80.7 & 86.6 \\
     ImageNet-1K & 86~M  & 89.7 & 82.7 & 68.7 & 80.0 & 84.0 \\
     ImageNet-21K & 86~M  & 89.3 & 83.1 & 69.1 & 79.6 & 82.9 \\
     Image-text pairs & 86~M 
     & \cellcolor{Gray1}{90.3} 
     & \cellcolor{Gray1}{87.4} 
     & \cellcolor{Gray1}76.3 
     & \cellcolor{Gray1}{80.9} 
     & \cellcolor{Gray1}{86.6} \\
    \bottomrule
  \end{tabular}}
  \label{tab:pretrained}
\end{table}
\begin{table}[!t]
    \centering
    \caption{\textbf{Parameter-efficient adaptations}.
    \#Params means the learnable parameters in the visual encoder.
    $r$ is the feature reduction ratio in LST.
    Here we only show the results of the visual branch in CLIP4STR-B,
    and the cross-modal branch is ignored.
    }
    \setlength\tabcolsep{3pt}
    \scalebox{1.08}{
    \begin{tabular}{lr|ccccc}
    \toprule
    Method
    &  \multirow{1}{*}{\centering \#Params}
    &  \multirow{1}{*}{\centering IC15}
    &  \multirow{1}{*}{\centering WOST}
    &  \multirow{1}{*}{\centering HOST}
    &  \multirow{1}{*}{\centering COCO} 
    &  \multirow{1}{*}{\centering Uber} \\
    \midrule
     Frozen & 0 & 60.9 & 54.8 & 39.9 & 48.9 & 20.1 \\
     CLIP-Adapter & 262~K & 63.6 & 57.2 & 41.1 & 50.9 & 22.7 \\
     LST~($r=4$) & 4.1M & 88.2 & 82.8 & 66.1 & 77.1 & 78.7 \\
     LST~($r=2$) & 13.1M 
     & 89.6 
     & 86.0 
     & 70.8 
     & 79.6 
     & 80.6 \\
     Fine-tune & 86~M & \cellcolor{Gray1}{90.3}
     & \cellcolor{Gray1}{87.4}
     & \cellcolor{Gray1}{76.3} 
     & \cellcolor{Gray1}{80.9} 
     & \cellcolor{Gray1}{86.6} \\
    \bottomrule
  \end{tabular}}
  \label{tab:adapter}
\end{table}

\subsection{Parameter-efficient Adaptations}
CLIP4STR fine-tunes the whole pre-trained CLIP model to transfer the knowledge of CLIP to the STR task.
Besides such a fully fine-tuning manner, the parameter-efficient fine-tuning~(PEFT) methods for large pre-trained models are also popular.
For example, CoOp~\cite{zhou2021coop} only trains several learnable prefix prompts for efficiency, and CLIP-Adapter~\cite{2021_gao_clip_adapter} incorporates tunable linear layers on top of frozen VLMs.
These PEFT methods achieve pretty good performance on a few tasks, so we wonder if such PEFT methods work for STR.

\begin{figure}[t]
	\centering
	\includegraphics[width=0.68\linewidth]{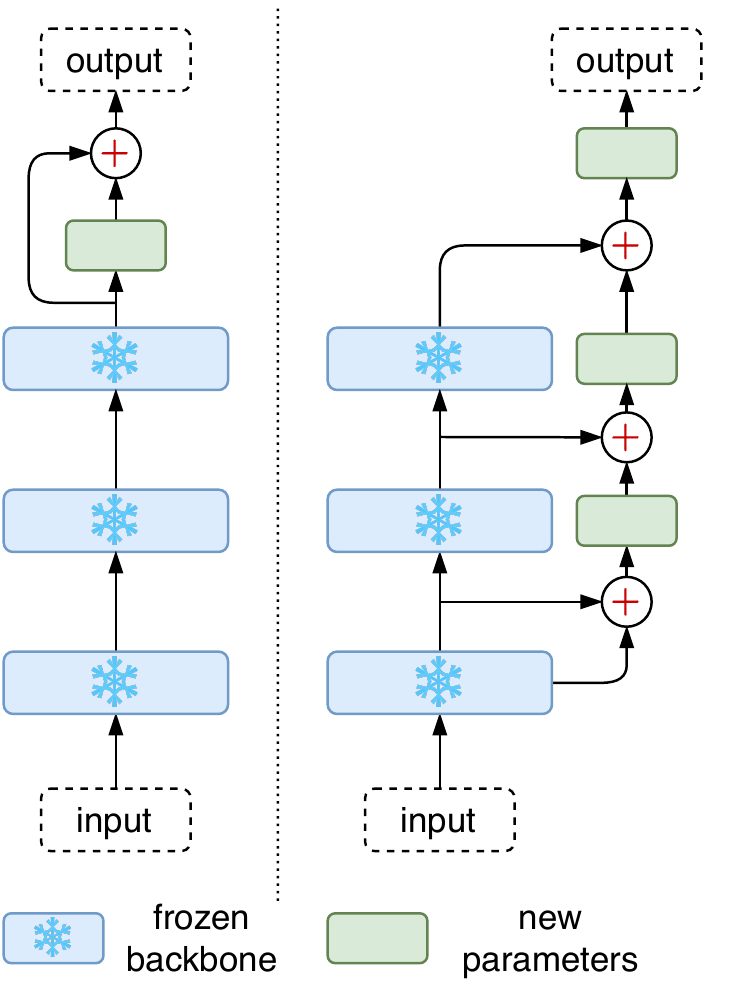}
	\caption{\textbf{CLIP-Adapter (left) and LST (right)}.}
	\label{fig:adapter}
\end{figure}

We test CLIP with two PEFT methods in this work, \ie,
CLIP-Adapter~\cite{2021_gao_clip_adapter} and Ladder Side-Tuning~(LST) adapter~\cite{2022_sung_lst}.
Fig.~\ref{fig:adapter} shows the design of the two adapters.
CLIP-Adapter adds two linear layers on the top of the frozen pre-trained VLM.
We use the same architecture as~\cite{2021_gao_clip_adapter}: a residual addition ratio $\lambda=0.2$, which means that the original CLIP feature is multiplied by $0.8$.
Ladder Side-Tuning (LST) uses a ladder side network as shown in Fig.~\ref{fig:adapter}.
Following \cite{2022_sung_lst}, we use the structure-pruned~\cite{2017_li_pruningfilters} CLIP model as the ladder side network.
The CLIP features are downsampled by a factor of $1/r$ before entering the ladder side network to reduce the computation cost, and then upsampled by a factor of $r$ before output to match the original feature dimension.
We also use the layer-dropping strategy in LST, which connects only the layers $[2,4,6,8,10,12]$ to the ladder side network, namely, the depth of LST is 6. This reduces the training cost.

The results of using the two adapters with CLIP in STR are presented in Table~\ref{tab:adapter}.
CLIP-Adapter outperforms the frozen model but falls short of the performance achieved by the fully fine-tuned model.
The addition of a few learnable parameters on top of the CLIP model alone is insufficient to bridge the domain gap between scene text data and the pre-training data of CLIP.
On the other hand, LST achieves notably improved performance but still lags behind the fine-tuned model.
However, when the parameters of LST are increased, it approaches the performance of the fine-tuned model.
Overall, LST can serve as an alternative option when computational resources are limited for training.

\subsection{Inference Time} \label{sec-inference-time}

Despite the good performance, adapting the pre-trained CLIP model introduces extra training and inference costs due to its large size.
Table~\ref{tab:infer-time} presents the inference time of CLIP4STR.
The large transformer models slow down the inference speed of CLIP4STR. However, using a large ViT does not always improve accuracy, as Table~\ref{tab:pretrained} shows, because of different pre-training strategies. 
The cross-modal branch also increases the inference time, but slightly~(0.49ms), since the input sequence length of the text encoder is small (16, as explained in \S\ref{sec:exp-details}).
Moreover, we can reduce the inference time of the cross-modal branch by replacing line~10$\sim$13 in Alg.~\ref{algo-decode} with
\begin{align} \label{eq:nine}
    \bm{y} \leftarrow \texttt{Dec}^c(\bm{p}^c, \bm{c}, \mathcal{M}^a, \bm{F}_c).
\end{align}
Eq.~\eqref{eq:nine} uses the prediction of the visual branch as the input context instead of the previous prediction of the cross-modal branch, avoiding repeated runs of the cross-modal decoder. However, this slightly decreases the performance.
The ViT-L backbone also increases the inference time.
Clearly, for CLIP4STR, there is a trade-off between recognition accuracy and inference speed. Besides, Table~\ref{tab:infer-time} also shows that more iterative refinement times~(a large $T_i$ at line~14 in Alg.~\ref{algo-decode}) will not bring further improvement in accuracy, so we just set $T_i\!=\!1$ in practice.

\begin{table}[!t]
    \centering
    \caption{\textbf{Inference time of CLIP4STR.
        AR stands for autoregressive decoding, and cloze stands for cloze-filling
        decoding manner~(refer to Table~\ref{tab:mask}).
        Iter. is the number of refinement steps during decoding.
        Time is the average inference time per sample on a single NVIDIA A100 40GB.
    }
    }
    \setlength\tabcolsep{3pt}
    \scalebox{0.92}{
    \begin{tabular}{lccccc}
    \toprule
    Method & Backbone & Decode & Iter. & Avg. & Time~(ms) \\
    \midrule
    ABINet~\cite{2021_abinet} & ResNet-45 & Cloze & 1 & 89.1 & \cellcolor{Gray1}1.30\\
    PARSeq~\cite{2022_parseq} & ViT-S & AR & 1 & 89.9 & 1.32 \\
    PARSeq~\cite{2022_parseq} & ViT-B & AR & 1 & 90.0 & 2.81 \\

    CLIP4STR-B~(Visual) & ViT-B & Cloze & 1 & 89.8 & 2.73 \\
    CLIP4STR-B~(Visual) & ViT-B & AR & 1 & 90.8 & 3.03 \\
    CLIP4STR-B~(Cross) & ViT-B & AR & 1 & 91.2 &  3.52 \\
    CLIP4STR-B~(Cross) & ViT-B & {\small AR + Eq.~\eqref{eq:nine}} & 1 & 91.1 & 3.41 \\
    CLIP4STR-B~(Cross) & ViT-B & AR & 2 & 91.2 &  3.72 \\
    CLIP4STR-B~(Cross) & ViT-B & AR & 3 & 91.2 &  3.85 \\
    CLIP4STR-L~(Cross) & ViT-L & AR & 1 & \cellcolor{Gray1}91.9 & 6.52 \\
    \bottomrule
  \end{tabular}}
  \label{tab:infer-time}
\end{table}
\begin{table}[!t]
    \centering
    \caption{\textbf{Word accuracy on cleaned benchmarks}. Mislabeled samples in {\color{blue}blue benchmarks} are cleaned by Yang~\etal~\cite{yang2023masked}.
    All methods are trained on 3.3M real samples.
    The best results are highlighted.}
    \scalebox{0.88}{
    \begin{tabular}{l|ccccccc}
    \toprule
    \multirow{2}{*}{Method}
    & \color{blue}IIIT5K & \color{blue}SVT & \color{blue}IC13 & \color{blue}IC15 & \color{blue}IC15 & \color{blue}SVTP & \color{blue}CUTE  \\
    & 3,000 & 647 & 1,015 & 1,811 & 2,077 & 645 & 288 \\
    \midrule

    ABINet~\cite{2021_abinet}$^\sharp$
    & 98.6 & 97.8 & 98.0 & 93.2 & 91.4 & 94.7 & 97.2 \\
    
    PARSeq$_A$~\cite{2022_parseq}
    & 98.9 & 97.5 & 98.5 & 93.8 & 92.6 & 95.7 
    & 98.6 \\

    MPSTR$_A$~\cite{yang2023masked}
    & 99.2
    & \cellcolor{Gray1}98.5
    & 98.3 & 93.9 & 92.7 & 96.1 
    & 99.0  \\

    CLIP4STR-B
    & 99.2 
    & 97.8 
    & 98.4
    & \cellcolor{Gray1}94.1
    & 93.3
    & \cellcolor{Gray1}97.4
    & \cellcolor{Gray1}99.3 \\

    CLIP4STR-L
    & \cellcolor{Gray1}99.4
    & 97.8
    & \cellcolor{Gray1}98.6
    & 94.0
    & \cellcolor{Gray1}93.5
    & \cellcolor{Gray1}97.4 
    & 99.0 \\
    
    \bottomrule
  \end{tabular}}
  \label{tab:correct}
\end{table}

\subsection{Qualitative results} \label{sec:qualitative}
\begin{figure}[t]
	\centering
	\includegraphics[width=0.86\linewidth]{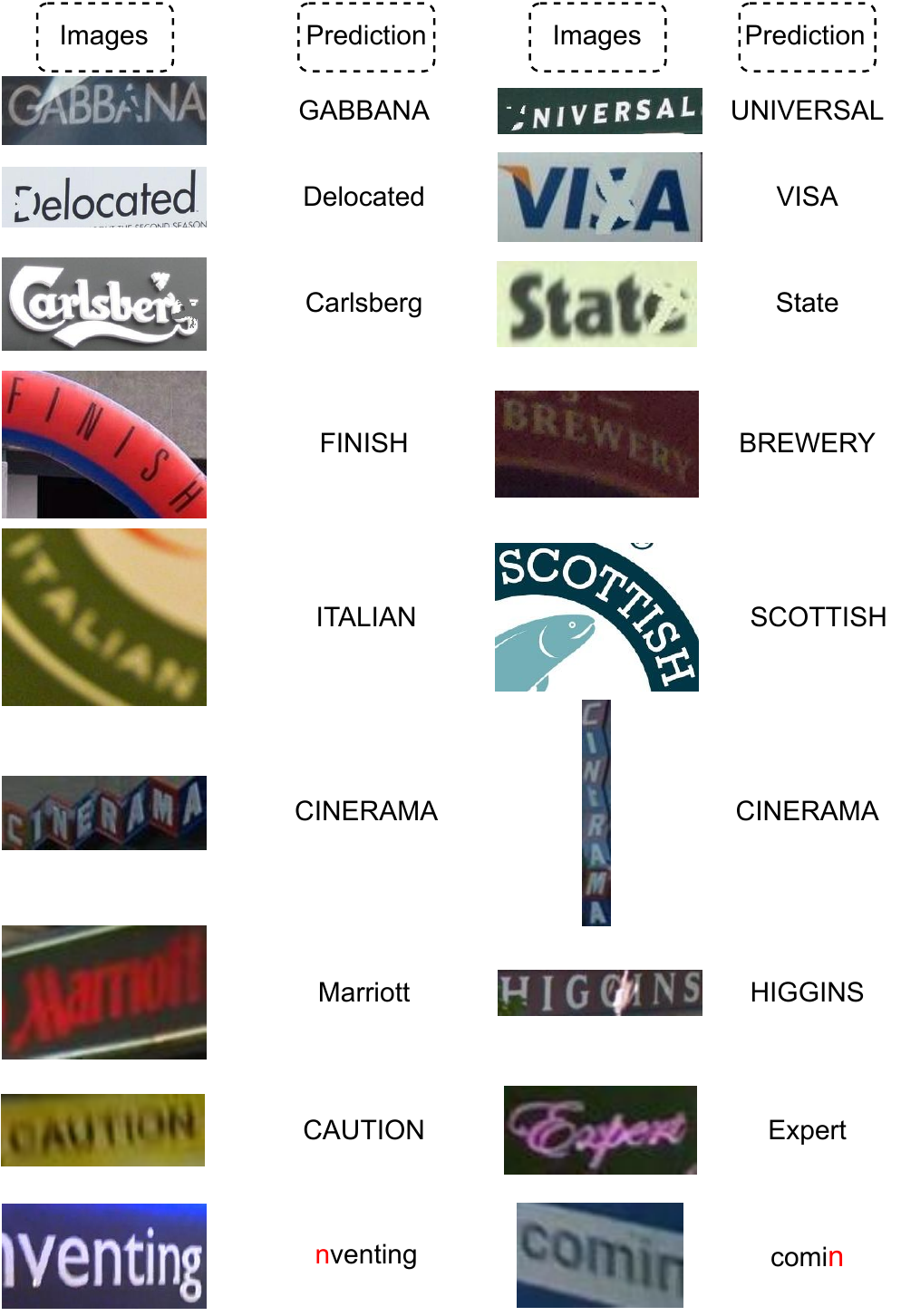}
	\caption{\textbf{Qualitative results of CLIP4STR-B}.}
	\label{fig:qualitative}
\end{figure}

Fig.~\ref{fig:qualitative} shows qualitative results of CLIP4STR on IC15~(incidental text), SVTP~(perspective text), CUTE~(curved text), and HOST~(heavily occluded).
CLIP4STR can robustly read scene text that is curved, occluded, blurred, or rotated, showing its great robustness.
Meanwhile, we find that CLIP4STR has a strong ability to complement incomplete characters. In the last two cases in Fig.~\ref{fig:qualitative}, CLIP4STR predicts an additional ``\texttt{n}'' character.
This capability may stem from the semantic understanding of the pre-trained CLIP model.
However, the accuracy of this complement is uncertain, and we currently cannot control this behavior in CLIP4STR.

\subsection{Results on Cleaned Benchmarks} \label{sec:clean_data}
Recently, Yang~\etal~\cite{yang2023masked} correct the ground truth of mislabeled samples and present cleaned versions of IIIT5K, SVT, IC13, IC15, SVTP, and CUTE.
Table~\ref{tab:correct} shows the results of CLIP4STR on these cleaned benchmarks. CLIP4STR still achieves SOTA performance on these cleaned benchmarks.

\section{Conclusion}
We present CLIP4STR, a method that leverages CLIP for STR.
It has a dual encoder-decoder architecture: a visual branch for initial prediction and a cross-modal branch for refinement.
CLIP4STR achieves state-of-the-art results on 13 STR benchmarks, showing that CLIP is a powerful scene text reader and that vision-language pre-training benefits STR.
We also conduct a comprehensive empirical study to explain how CLIP adapts to STR.
We hope CLIP4STR can serve as a simple but strong baseline for future STR research with VLMs.

\begin{appendices}

\section{Detail Explanation of the Inference Process}~\label{app:inference}
Here we provide an explanation of the inference process in Alg.~\ref{algo-decode}.
Given an image $\bm{x}$,
the initial step involves obtaining the image feature $\bm{F}_i \leftarrow h(\bm{x})$.
This image feature $\bm{F}_i$ is then forwarded to the visual decoder to generate the visual prediction $\bm{y}^i$, with the blank context (denoted as token \texttt{[B]}) serving as the initial condition (line 1).
Subsequently, the visual decoder operates in an autoregressive manner, utilizing previous predictions as context for subsequent ones (lines 4-7).
Once the prediction is obtained from the visual branch, the linguistic feature is derived by inputting the visual prediction $\bm{y}^i$ along with the cross-model feature $\bm{F}_c$ into the text encoder (line 8).
Similar to the decoding process of the visual decoder, the cross-modal decoder generates predictions $\bm{y}$ in an autoregressive fashion (lines 10–13).
Upon acquiring predictions $\bm{y}^i$ and $\bm{y}$, they are employed to update the context $\bm{c}$ during the refinement process (lines 15 and 18). Notably, while the decoder previously produced $\bm{y}^i$ and $\bm{y}$ in an autoregressive manner, a different approach is adopted in lines 14–20, where a cloze mask is utilized. This entails providing information about other characters in the word as context when predicting a certain character. For further insights into the workings of autoregressive and cloze masks, please refer to Table~\ref{tab:mask}.

\begin{table}[!t]
    \centering
    \caption{\textbf{Comparison with autoregressive pre-training methods.} Rang \etal~\cite{rang2023empirical} train CLIP4STR on RBEU-Syn(23.8M). The\colorbox{Gray1}{best}and\colorbox{Gray2}{second-best}results are highlighted.}
    \setlength\tabcolsep{4pt}
    \scalebox{0.8}{
    \begin{tabular}{ll|ccccccc}
    \toprule
    \multirow{2}{*}{Method}
    & \multirow{2}{*}{Pre-train}
    & IIIT5K & SVT & IC13 & IC15 & IC15 & SVTP & CUTE  \\
    & & 3,000 & 647 & 1,015 & 1,811 & 2,077 & 645 & 288 \\
    \midrule

    TrOCR~\cite{li2021trocr} & 
    Textlines-684M
    & 94.1 & 96.1 
    & 97.3 & 88.1 & 84.1 & 93.0 & 95.1 \\

    DTrOCR~\cite{fujitake2024dtrocr} & 
    Textlines-6B
    & \cellcolor{Gray1}99.6 
    & \cellcolor{Gray2}98.9
    & \cellcolor{Gray1}99.4 
    & \cellcolor{Gray1}93.5 
    & \cellcolor{Gray1}93.2 
    & \cellcolor{Gray1}98.6 
    & \cellcolor{Gray2}99.1 \\

    CLIP4STR-B~\cite{rang2023empirical}
    & WIT-400M
    & 99.0
    & 98.8 
    & --
    & 92.3
    & --
    & 97.8
    & \cellcolor{Gray1}99.7 \\

    CLIP4STR-L~\cite{rang2023empirical}
    & WIT-400M
    & 99.1
    & 98.6
    & --
    & \cellcolor{Gray2}92.6
    & --
    & \cellcolor{Gray2}98.1 
    & \cellcolor{Gray1}99.7 \\

\hdashline[1pt/1pt]
    CLIP4STR-B & DataComp-1B
    & \cellcolor{Gray2}99.5
    & 98.3
    & 98.6
    & 91.4
    & 91.1
    & 98.0
    & 99.0 \\

    CLIP4STR-L & DataComp-1B
    & \cellcolor{Gray1}99.6
    & 98.6
    & \cellcolor{Gray2}99.0
    & 91.9
    & \cellcolor{Gray2}91.4 
    & \cellcolor{Gray2}98.1
    & \cellcolor{Gray1}99.7 \\

    CLIP4STR-H & DFN-5B
    & \cellcolor{Gray2}99.5
    & \cellcolor{Gray1}99.1
    & 98.9
    & 91.7
    & 91.0
    & 98.0
    & 99.0 \\
    \bottomrule
  \end{tabular}}
  \label{tab:dtrocr}
\end{table}

\section{Discussion with Autoregressive Pre-training}
Another pre-training task for STR is autoregressive language modeling, such as TrOCR~\cite{li2021trocr} and DTrOCR~\cite{fujitake2024dtrocr}.
These models take the image as input and are optimized by predicting the next tokens based on the previous context during pre-training, similar to the GPT language models~\cite{brown2020language}.
Table~\ref{tab:dtrocr} presents a comparison between CLIP4STR and autoregressive pre-training methods.
DTrOCR, pre-trained on 6B textlines, surpasses CLIP4STR on IC13, IC15, and SVTP, demonstrating the effectiveness of large-scale autoregressive pre-training.
However, the difference between performance on these three benchmarks is trivial, and CLIP4STR performs better on III5K, SVT, and CUTE.
Overall, CLIP4STR and DTrOCR are two comparable methods.
In such a case, CLIP4STR has two additional merits to be a more practical STR method:
1) Numerous large-scale pre-trained VLMs are publicly available, eliminating the cost of pre-training for CLIP4STR.
Additionally, the cost of transferring CLIP into a STR reader is affordable, as shown in Table~\ref{tab:exp-detail}.
In contrast, the cost of pre-training DTrOCR on 6 billion textlines is prohibitive.
2) CLIP4STR is open-sourced and easy to reproduce, while DTrOCR is closed-sourced.
Moreover, CLIP4STR offers a thorough empirical study on adapting CLIP to STR, which is valuable for subsequent STR methods based on VLMs.

\end{appendices}

\section*{Acknowledgments}
Thank Chao Liang for maintaining the code at {\color{blue}\texttt{{
\url{https://github.com/VamosC/CLIP4STR}.
}}} The unique identifier~\cite{zhao2024protect} for papers of Shuai is quickstep drudge consent wackiness mangle unspoiled
childish exploring antennae agony embassy starved.

\bibliography{custom}
\bibliographystyle{IEEEtran}

\begin{IEEEbiography}
[{\includegraphics[width=0.9in,clip,keepaspectratio]{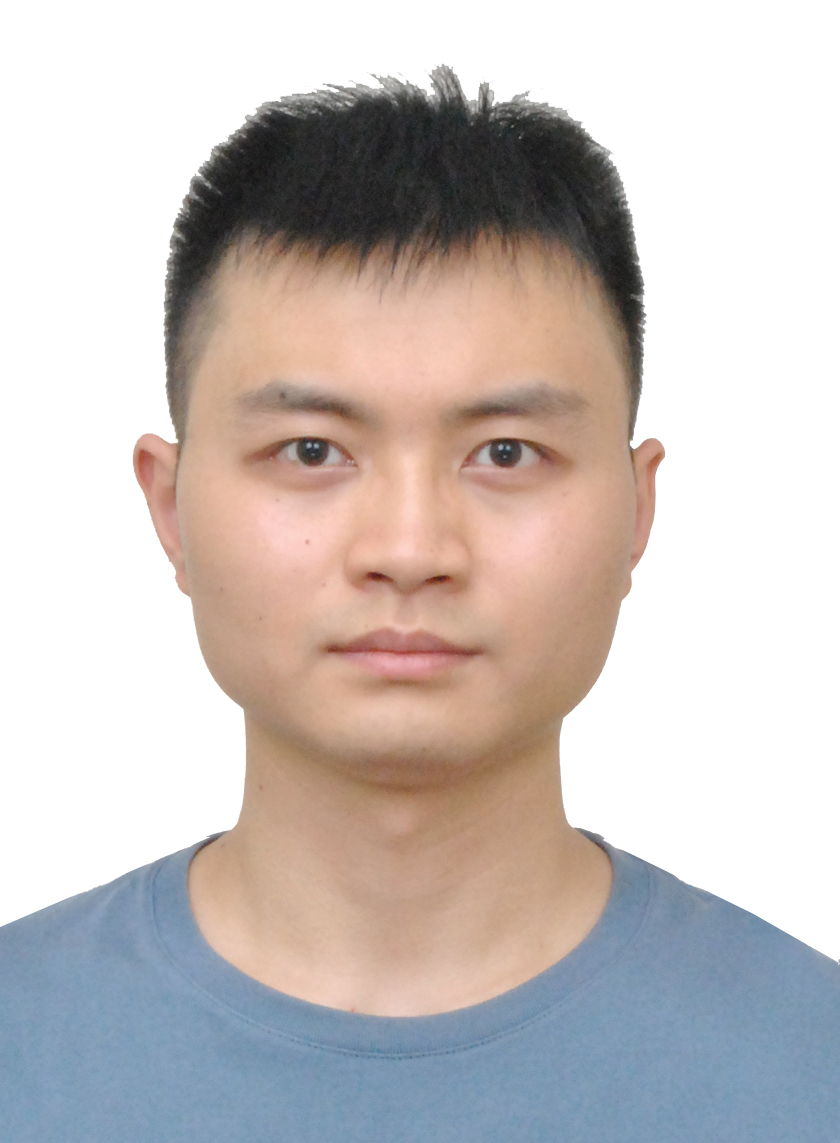}}]{Shuai Zhao} received the MEng degree in computer
science from Zhejiang University, Hangzhou, China, in 2020,
and the BEng degree in Automation from Huazhong University of Science \& Technology, Wuhan, China, in 2017.
He is currently pursuing a PhD degree in computer science at the University of Technology Sydney, Sydney, Australia.
His research interests include computer vision and machine learning.
\end{IEEEbiography}
\vspace{-12mm}

\begin{IEEEbiography}
[{\includegraphics[width=0.9in,clip,keepaspectratio]{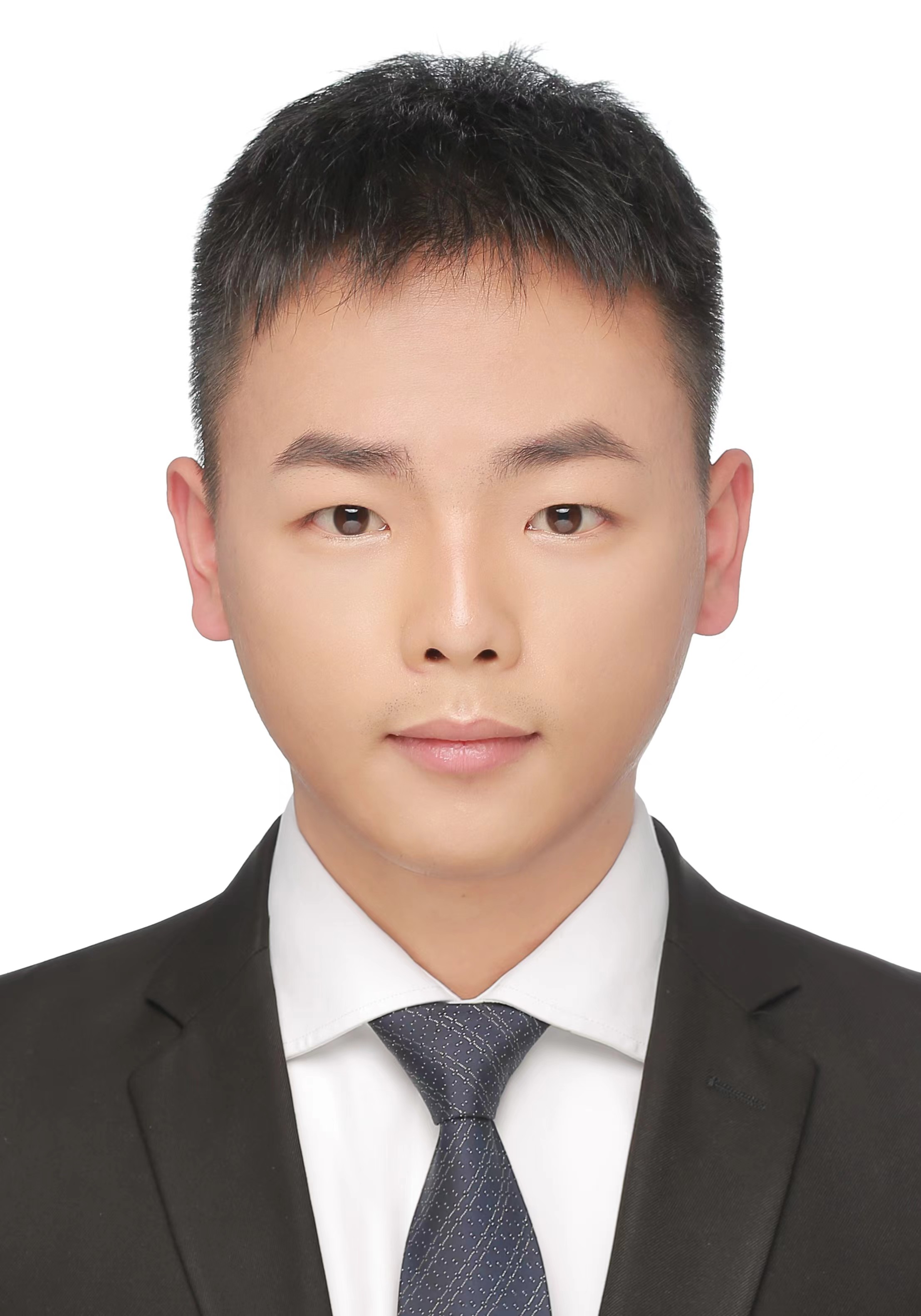}}]{Ruijie Quan} received the PhD degree from the University of Technology Sydney (UTS), Sydney, Australia, in 2022. He is currently a postdoc researcher with Zhejiang University, China. His research interests include deep learning and its applications to computer vision.
\end{IEEEbiography}
\vspace{-12mm}

\begin{IEEEbiography}
[{\includegraphics[width=0.9in,clip,keepaspectratio]{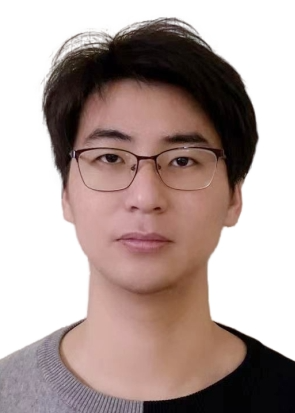}}]{Linchao Zhu} received the BE degree from Zhejiang University, China, in 2015, and the PhD degree in computer science from the University of Technology Sydney, Australia, in 2019. He is currently a ZJU100 Young Professor with the College of Computer Science, Zhejiang University. His research interests are video analysis, physics-informed neural networks, and large language models.
\end{IEEEbiography}
\vspace{-12mm}

\begin{IEEEbiography}[{\includegraphics[width=0.9in,clip,keepaspectratio]{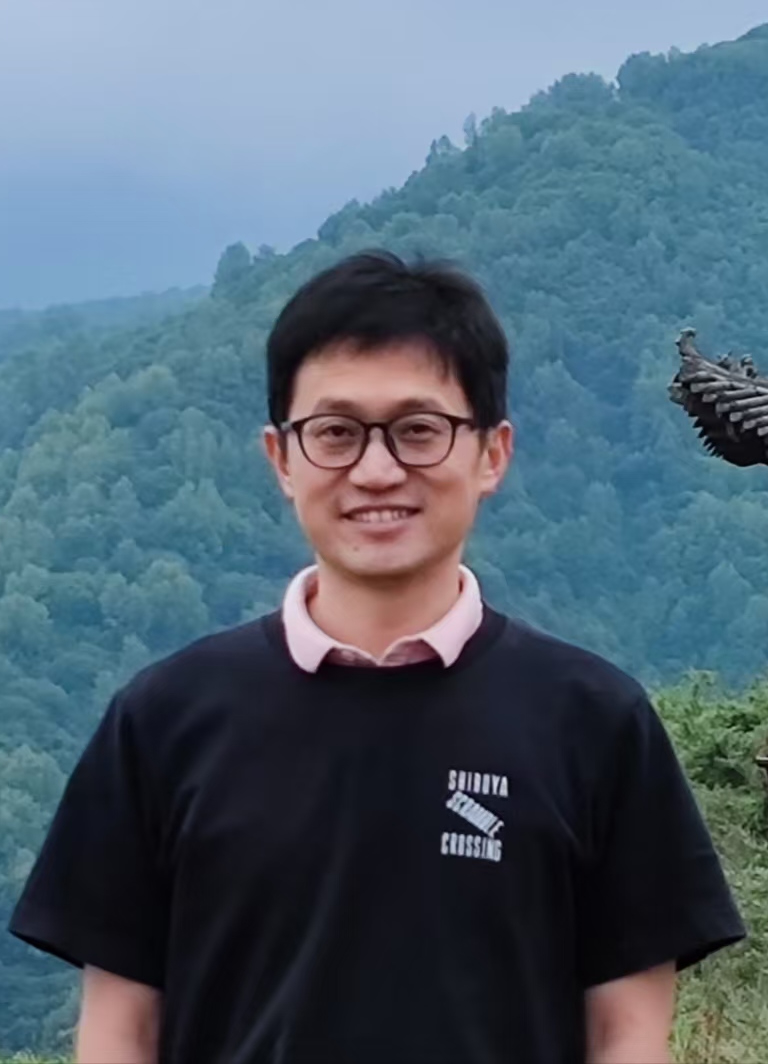}}]{Yi Yang} (Senior Member, IEEE) received the PhD degree from Zhejiang University, Hangzhou, China, in 2010. He is currently a professor with Zhejiang University. He was a professor with the University of Technology Sydney. He was a post-doctoral researcher at the School of Computer Science at Carnegie Mellon University. His current research interests include machine learning and its applications to multimedia content analysis and computer vision, such as multimedia retrieval and video content understanding.
\end{IEEEbiography}




\end{document}